\documentclass{article}

\PassOptionsToPackage{numbers, compress}{natbib}
 \usepackage[main, final]{neurips_2026}
\usepackage[utf8]{inputenc} % allow utf-8 input
\usepackage[T1]{fontenc}    % use 8-bit T1 fonts
\usepackage{url}            % simple URL typesetting
\usepackage{booktabs}       % professional-quality tables
\usepackage{amsfonts}       % blackboard math symbols
\usepackage{nicefrac}       % compact symbols for 1/2, etc.
\usepackage{microtype}      % microtypography
\usepackage{xcolor}         % colors
\usepackage{amsmath}
\usepackage{multirow}
\usepackage{graphicx}
\usepackage[toc,page]{appendix}
\usepackage{makecell}
\usepackage{enumitem} 
\usepackage{subcaption}
\usepackage{algorithm}
\usepackage{algpseudocode}

\usepackage{amsthm}

\usepackage{hyperref}       % hyperlinks

\newtheorem{assumption}{Assumption}

\newtheorem{proposition}{Proposition}

\newtheorem{theorem}{Theorem}

\newcommand{\gy}[1]{\textcolor{gray}{#1}}          
\newcommand{\bd}[1]{\textbf{#1}}                  
\newcommand{\bb}[1]{\textbf{\textcolor{blue}{#1}}} 
          
\newcommand{\sd}[1]{{\scriptsize\,$\pm$#1}}
\title{Safe Offline Reinforcement Learning using Behavior Regularisation and Latent Feasibility Guidance}

\author{%
  Mahesh Keswani \\
  Yardi School of Artificial Intelligence \\
  IIT Delhi \\
  \texttt{aiy247544@iitd.ac.in} \\
  \And
  Parth Bhardwaj \\
  Dept.\ of Mathematics and Computing \\
  IIT Delhi \\
  \texttt{mt1221257@maths.iitd.ac.in} \\
  \AND
  Ankur Kumar \\
  Yardi School of Artificial Intelligence \\
  IIT Delhi \\
  \texttt{aib252557@scai.iitd.ac.in} \\
  \And
  Raunak P. Bhattacharyya \\
  Yardi School of Artificial Intelligence \\
  IIT Delhi \\
  \texttt{raunakbh@iitd.ac.in}
}
\begin{document}

\maketitle

\begin{abstract}
Safe offline reinforcement learning (RL) seeks to learn policies that maximize return while
satisfying safety constraints, using a static dataset. To characterize
state-wise safety, reachability analysis identifies the feasible region, defined as the
set of states from which constraints can be satisfied at every future step. Policies are commonly extracted from learned value functions with
advantage-weighted regression (AWR). However, AWR relies on exponentiated weights that amplify value estimation
errors, whereas behavior-regularized policy gradient methods directly utilize the action
gradient of the value function and often make more effective use of learned critics. We
propose Feasibility-Optimized Conditional Actor Learning (FOCAL), a method that incorporates
behavior-regularized policy gradients into a conditional variational autoencoder
(CVAE) actor. FOCAL optimizes this policy using a first-order objective partitioned by
feasibility, updating the actor to increase return on transitions estimated to be feasible
and to reduce constraint violations on transitions estimated to be infeasible. During inference,
FOCAL dynamically adjusts the scale of the latent prior according to the estimated feasibility
of the state to adjust the conservatism of action sampling. Theoretically, we show that for AWR,
the policy update variance depends exponentially on the advantage error, whereas for FOCAL
it scales linearly with the variance of the value estimation error. Across the DSRL benchmark,
FOCAL satisfies the constraints on all evaluated tasks while achieving competitive
or higher returns compared to baselines.
\end{abstract}

\section{Introduction}
\label{sec:introduction}

Decision-making systems deployed in safety-critical domains, such as robotics, autonomous
driving, and industrial control, must satisfy explicit safety constraints, since constraint
violations can have severe consequences \cite{kiran2021deep, zhan2022deepthermal}. Safe
reinforcement learning (RL) formalizes this requirement by optimizing task performance
subject to constraints \cite{altman2021constrained}. Learning such policies through online
interaction, however, requires exploration, which inherently risks violating constraints
during training. Safe offline RL avoids this risk by learning policies from previously
collected datasets without environment interaction \cite{levine2020offline,
xu2022constraints}. A safe offline RL method must therefore balance three interconnected
objectives: it must improve task performance, satisfy safety constraints, and remain close
to the behavioral data distribution to limit out-of-distribution (OOD) value extrapolation
errors \cite{xu2022constraints, zheng2024safe}.

Two principal design choices determine how safe offline RL methods address these
objectives. The first is the representation of the safety constraint. A common approach is
to constrain the expected cumulative cost \cite{xu2022constraints, lee2022coptidice,
liu2023constrained}, which bounds aggregate violations across trajectories but does not
explicitly address violations at individual states. An alternative is to reason about
state-wise feasibility. A state is called feasible if there exists a policy that, starting
from this state, satisfies the constraint at every future step, and infeasible if all
policies eventually violate the constraint. Hamilton-Jacobi (HJ) reachability characterizes
the feasible set through a value function that measures the worst-case future constraint
violation \cite{bansal2017hamilton, fisac2019bridging, yu2022reachability}. We refer to a
learned estimate of this function, evaluated for a state-action pair, as the hazard critic.
\citet{zheng2024safe} showed that the hazard critic can be estimated from offline data and
used to assign distinct policy extraction objectives based on feasibility: maximizing the
return in feasible states, and minimizing future violations in infeasible states. The second
design choice is the extraction of the policy from the learned value functions. Methods that
decouple critic learning from policy learning commonly extract the policy with
advantage-weighted regression (AWR) \cite{peng2019advantage, kostrikov2021offline,
zheng2024safe, koirala2024latent}, which weights the behavior-cloning likelihood of each
dataset action by its exponentiated estimated advantage. Since the critics are learned from
finite data, the estimated advantages contain errors, and in AWR these errors enter the
policy update through the exponential weight. As a result, the update can be dominated by a
small number of transitions with large estimated advantages \cite{park2024value}. \citet{park2024value} further showed that the extraction step can affect performance as much
as value learning, and that behavior-regularized policy gradient methods
\cite{fujimoto2021minimalist} often make more effective use of learned critics than weighted
regression.

Policy extraction in safe offline RL is subject to additional challenges. First, the policy
is guided by both a reward critic and a hazard critic, which act on the same action.
Combining their gradients in a single objective requires a trade-off between the two, as in
Lagrangian formulations, whereas a feasibility-dependent formulation assigns a single
objective to each transition according to the estimated feasibility of its state. Second,
the policy must represent the behavior distribution of the dataset while supporting
efficient inference. Diffusion-based policies are expressive but require iterative denoising
steps that increase inference latency \cite{hansen2023idql,
zheng2024safe}, whereas latent-variable models such as conditional variational autoencoders
(CVAEs) generate actions in a single forward pass. Third, offline policies encounter
distribution shift at deployment \cite{park2024value}. In the safe setting, the degree
of conservatism of the policy should therefore depend on the estimated feasibility of the
current state.

To address these challenges, we propose Feasibility-Optimized Conditional Actor Learning
(FOCAL), a safe offline RL method that extracts a single-step generative policy from learned
reward and hazard critics. FOCAL first estimates both critics from the offline data. It then
trains a CVAE actor with a feasibility-partitioned first-order objective, a
behavior-regularized policy gradient objective in which the critic that guides each
transition is selected according to the estimated feasibility of its state. For transitions
in the estimated feasible region, the actor is updated in the direction that increases the
estimated return, and for transitions in the estimated infeasible region, it is updated in
the direction that decreases the estimated constraint violation. In both cases, a
reconstruction term keeps the generated actions close to the dataset actions. Each
transition therefore receives the gradient of exactly one critic, and no trade-off weight
between reward and constraint is required. At deployment, FOCAL sets the scale of the latent
prior according to the estimated feasibility of the current state. Latent samples are drawn with the standard prior scale in states estimated to be far from a violation and are concentrated in
the high-probability region of the prior in states estimated to be close to a violation. The
resulting candidate actions are then filtered using the learned critics.

We analyze FOCAL theoretically and empirically. Theoretically, we characterize how
critic estimation errors enter the policy update. For AWR, the variance of the update depends
exponentially on the advantage
error, whereas for FOCAL the variance of the update depends linearly on the variance of the
critic error. We further bound the deviation of the training target from the dataset action
and the value of the hazard critic at this target. Empirically, evaluations on the Datasets
for Safe Reinforcement Learning (DSRL) \cite{liu2023datasets} show that FOCAL
satisfies the specified cost limits across all evaluated tasks while achieving competitive
or higher returns compared to baselines.

In summary, our main contributions are:
\begin{itemize}
    \item We propose a feasibility-partitioned first-order objective for optimizing a
single-step CVAE actor, which enables behavior-regularized policy gradient in safe
offline RL.
    \item We introduce an adaptive inference procedure that sets the scale of the latent
    prior based on the estimated state-wise feasibility.
    \item We analyze how critic estimation error enters the policy update under
    advantage-weighted regression and under the proposed objective, bound the deviation of
    the training target from the dataset action, and evaluate FOCAL on the DSRL benchmark.
\end{itemize}
\section{Preliminaries}
\label{sec:prelims}

\subsection{Safe Offline Reinforcement Learning}

We formulate safe reinforcement learning as a Constrained Markov Decision Process (CMDP),
defined by the tuple $\mathcal{M} = (\mathcal{S}, \mathcal{A}, P, r, c, h, \gamma, \rho_0)$
\cite{altman2021constrained, zheng2024safe}. Here, $\mathcal{S}$ and $\mathcal{A}$ denote
the state and action spaces, $P(s' \mid s,a)$ denotes the transition dynamics, $r(s,a)$ is
the reward function, $c: \mathcal{S} \to \mathbb{R}_{\ge 0}$ is the cost function,
$\gamma \in (0,1)$ is the discount factor, and $\rho_0$ is the initial state distribution.
The constraint violation function $h: \mathcal{S} \to \mathbb{R}$ characterizes state-wise
safety: a state $s$ satisfies the safety constraint if $h(s) \le 0$, and $h(s) > 0$ denotes
a constraint violation. In this work, $h$ is constructed from the cost function as described
in Section~\ref{sec:feasibility_identification}.

In the offline setting, the agent learns a policy $\pi(a \mid s)$ from a static dataset
$\mathcal{D} = \{(s,a,s',r,c)\}$ collected by an unknown behavior policy $\pi_\beta$, which
contains both safe and unsafe trajectories. Expectation-based safe offline RL methods
optimize the policy under an expected cost limit and a behavior regularization constraint
\cite{xu2022constraints, lee2022coptidice, zheng2024safe}:
\begin{equation}
\label{eq:soft_constraint}
\max_{\pi}\;
\mathbb{E}_{s \sim \rho_0}\big[ V_r^\pi(s) \big]
\quad
\text{s.t.}\quad
\mathbb{E}_{s \sim \rho_0}\big[ V_c^\pi(s) \big] \le \kappa,
\qquad
D(\pi \,\Vert\, \pi_\beta) \le \epsilon,
\end{equation}
where
$V_r^\pi(s) = \mathbb{E}_{\pi}\big[ \sum_{t=0}^{\infty} \gamma^t r(s_t, a_t) \mid s_0 = s \big]$
is the reward state-value function,
$V_c^\pi(s) = \mathbb{E}_{\pi}\big[ \sum_{t=0}^{\infty} \gamma^t c(s_t) \mid s_0 = s \big]$
is the cost state-value function, $\kappa \ge 0$ is the cost limit, and
$D(\pi \,\Vert\, \pi_\beta)$ is a divergence that limits the
distribution shift from the behavior policy, with tolerance $\epsilon > 0$.

\subsection{Feasible Region Identification}

To characterize state-wise safety, we use Hamilton-Jacobi (HJ) reachability
\cite{bansal2017hamilton, fisac2019bridging, yu2022reachability, zheng2024safe}. HJ reachability characterizes the set of feasible states through a
value function that measures the worst-case constraint violation along future trajectories,
minimized over policies. Following \citet{yu2022reachability} and \citet{zheng2024safe}, let $V_h^\pi(s)$ denote the maximum constraint violation
along the trajectory induced by policy $\pi$ from state $s$, and let $Q_h^\pi(s,a)$ denote the
corresponding quantity when the action is fixed to $a$. The optimal feasible state-value
and action-value functions are defined as
\begin{align}
V_h^*(s) &:= \min_{\pi} V_h^\pi(s) := \min_{\pi} \max_{t \in \mathbb{N}} h(s_t),
\qquad s_0 = s,\; a_t \sim \pi(\cdot \mid s_t),
\label{eq:vh_star}
\\
Q_h^*(s,a) &:= \min_{\pi} Q_h^\pi(s,a) := \min_{\pi} \max_{t \in \mathbb{N}} h(s_t),
\qquad s_0 = s,\; a_0 = a,\; a_{t+1} \sim \pi(\cdot \mid s_{t+1}).
\label{eq:qh_star}
\end{align}
A state $s$ is feasible if $V_h^*(s) \le 0$, that is, if there exists a policy under which
the constraint is satisfied at every step of the trajectory from $s$. The largest feasible
set is
\begin{equation}
\label{eq:feasible_set}
\mathcal{S}_f^* := \big\{\, s \in \mathcal{S} \;\big|\; V_h^*(s) \le 0 \,\big\}.
\end{equation}

This characterization motivates a feasibility-dependent formulation
\cite{yu2022reachability, zheng2024safe}. Within the feasible set $\mathcal{S}_f^*$, the
objective is to maximize the expected return subject to the state-wise constraint. In
infeasible states, the objective is to minimize the magnitude of future violations:

\noindent
\begin{minipage}[t]{0.48\textwidth}
\begin{equation}
\begin{aligned}
\textbf{Feasible:} \quad &
\max_{\pi}\;
\mathbb{E}_{s}\big[ V_r^\pi(s)\, \mathbb{I}_{s \in \mathcal{S}_f^*} \big] \\
\text{s.t.} \quad &
V_h^\pi(s) \le 0, \quad \forall s \in \mathcal{S}_f^*, \\
& D_{\mathrm{KL}}(\pi \,\Vert\, \pi_\beta) \le \epsilon,
\end{aligned}
\label{eq:coupled_feas}
\end{equation}
\end{minipage}%
\hfill%
\begin{minipage}[t]{0.48\textwidth}
\begin{equation}
\begin{aligned}
\textbf{Infeasible:} \quad &
\max_{\pi}\;
\mathbb{E}_{s}\big[ -V_h^\pi(s)\, \mathbb{I}_{s \notin \mathcal{S}_f^*} \big] \\
\text{s.t.} \quad &
D_{\mathrm{KL}}(\pi \,\Vert\, \pi_\beta) \le \epsilon,
\end{aligned}
\label{eq:coupled_infeas}
\end{equation}
\end{minipage}

\noindent
where $\mathbb{E}_s$ denotes expectation over dataset states. Directly optimizing
Eqs.~(\ref{eq:coupled_feas}) and~(\ref{eq:coupled_infeas}) requires training the policy
jointly with the value functions $V_r^\pi$ and $V_h^\pi$, which can lead to instability
\cite{lee2022coptidice, zheng2024safe}.

\subsection{Implicit Q-Learning for Reward Estimation}
\label{sec:iql_reward}

To estimate the reward value functions without querying OOD actions,
we employ Implicit Q-Learning (IQL) \cite{kostrikov2021offline}. IQL trains a state-value
network $\hat V_r$ and action-value networks $\hat Q_{r,i}$, $i \in \{1,2\}$, using only
actions contained in the dataset. $\hat V_r(s)$ is trained to approximate an upper expectile
of $\hat Q_r(s,a)$ over the dataset actions at state $s$, where
$\hat Q_r(s,a) = \min_{i=1,2} \hat Q_{r,i}(s,a)$ (Eq.~(\ref{eq:twin_critics})), and each
$\hat Q_{r,i}$ is trained by temporal difference learning:
\begin{align}
\mathcal{L}(\hat V_r)
&=
\mathbb{E}_{(s,a)\sim\mathcal{D}}
\Big[ L^{\omega_r}\big( \hat Q_r(s,a) - \hat V_r(s) \big) \Big],
\label{eq:vr_loss}
\\
\mathcal{L}(\hat Q_{r,i})
&=
\mathbb{E}_{(s,a,s')\sim\mathcal{D}}
\Big[ \big( r(s,a) + \gamma \hat V_r(s') - \hat Q_{r,i}(s,a) \big)^2 \Big],
\qquad i \in \{1,2\},
\label{eq:qr_loss}
\end{align}
where the asymmetric loss is
\begin{equation}
\label{eq:expectile_loss}
L^{\omega_r}(u) = \lvert \omega_r - \mathbb{I}(u < 0) \rvert\, u^2,
\qquad \omega_r \in (0.5, 1).
\end{equation}
As $\omega_r \rightarrow 1$, $\hat V_r(s)$ approaches the maximum of $\hat Q_r(s,a)$ over
actions in the support of $\pi_{\beta}$ \cite{kostrikov2021offline}:
\begin{equation}
\label{eq:expectile_max}
\lim_{\omega_r \rightarrow 1} \hat V_r(s)
=
\max_{a \in \mathcal{A}:\, \pi_\beta(a \mid s) > 0} \hat Q_r(s,a).
\end{equation}
The estimated reward advantage is $\hat A_r(s,a) = \hat Q_r(s,a) - \hat V_r(s)$.

\subsection{Conditional Variational Autoencoder}
\label{sec:cvae_prelim}

A Conditional Variational Autoencoder (CVAE) is a generative model that reconstructs data
conditioned on information \cite{doersch2016tutorial}. In RL, a CVAE can
model the conditional distribution of actions given the state, which enables single-step
policy inference. A CVAE consists of an encoder $q_\phi(z \mid s,a)$ and a decoder
$p_\theta(a \mid s,z)$, and is trained by maximizing the evidence lower bound (ELBO):
\begin{equation}
\label{eq:elbo}
\mathcal{L}_{\mathrm{ELBO}}(\theta,\phi)
=
\mathbb{E}_{(s,a)\sim\mathcal{D}}
\Big[
\mathbb{E}_{z \sim q_\phi(\cdot \mid s,a)}\big[ \log p_\theta(a \mid s,z) \big]
- \beta\, D_{\mathrm{KL}}\big( q_\phi(z \mid s,a) \,\Vert\, p(z) \big)
\Big],
\end{equation}
where $p(z) = \mathcal{N}(0,I)$ is the prior and $\beta > 0$ controls the strength of the
latent regularization.
\section{Methodology}
\label{sec:method}

This section presents FOCAL. Section~\ref{sec:feasibility_identification} describes the offline estimation of the
feasibility value functions. Section~\ref{sec:actor_learning} introduces the
feasibility-partitioned first-order objective used for policy extraction.
Section~\ref{sec:adaptive_inference} describes the adaptive test-time inference
procedure, and Section~\ref{sec:theory} states the theoretical properties of the
objective.

\subsection{Offline Feasibility Identification}
\label{sec:feasibility_identification}

FOCAL decouples critic training from policy optimization in order to avoid the
training instability that can arise from jointly optimizing the reward, feasibility,
and policy objectives in Eqs.~(\ref{eq:coupled_feas}) and~(\ref{eq:coupled_infeas})
\cite{lee2022coptidice, zheng2024safe}. The reward critics $(\hat{V}_r, \hat{Q}_r)$ are trained with
IQL as described in Section~\ref{sec:iql_reward}, and the
feasibility value functions $(\hat{V}_h, \hat{Q}_h)$ are estimated independently from the static
dataset. Throughout, hats denote learned critics, and starred quantities denote the
corresponding exact value functions.

\textbf{Constraint violation function.} Following \citet{zheng2024safe}, we construct the state-wise constraint violation
function from the per-step cost as
\begin{equation}
\label{eq:h_definition}
h(s) =
\begin{cases}
-1, & \text{if } c(s) = 0,\\
M, & \text{if } c(s) > 0,
\end{cases}
\end{equation}
with $M = 25$. A state is classified as feasible if
$\hat{V}_h(s) \le 0$ and as infeasible if $\hat{V}_h(s) > 0$ \cite{zheng2024safe}. 

\textbf{Feasibility critics.}
Computing $Q_h^*$ exactly requires access to the transition dynamics and an evaluation
over future trajectories. We therefore approximate it offline by dynamic programming with
the feasible Bellman operator, whose fixed point approaches $Q_h^*$ as the discount
factor satisfies $\gamma \rightarrow 1$ \cite{yu2022reachability, zheng2024safe}. Applying
this operator requires $\hat{V}_h(s') = \min_{a'} \hat{Q}_h(s', a')$. Minimizing over the entire
action space queries OOD actions and leads to underestimation of
the hazard value in the offline setting. To restrict the minimization to actions
represented in the dataset without an explicit behavior model, we adopt reversed
expectile regression \cite{kostrikov2021offline, zheng2024safe}. To reduce the effect of
approximation error, we use twin hazard critics $\hat{Q}_{h,1}, \hat{Q}_{h,2}$ and twin reward
critics $\hat{Q}_{r,1}, \hat{Q}_{r,2}$ \cite{fujimoto2018addressing}, and define the conservative
hazard estimate and the pessimistic reward estimate as
\begin{equation}
\label{eq:twin_critics}
\hat{Q}_h(s,a) = \max_{i=1,2} \hat{Q}_{h,i}(s,a),
\qquad
\hat{Q}_r(s,a) = \min_{i=1,2} \hat{Q}_{r,i}(s,a).
\end{equation}
The hazard state-value function is trained with the asymmetric loss
$L_{\mathrm{rev}}^{\omega_h}(u) = \lvert \omega_h - \mathbb{I}(u > 0) \rvert\, u^2$,
with expectile parameter $\omega_h \in (0.5, 1)$:
\begin{equation}
\label{eq:vh_loss}
\mathcal{L}(\hat{V}_h)
=
\mathbb{E}_{(s,a)\sim\mathcal{D}}
\Big[
L_{\mathrm{rev}}^{\omega_h}\big(\hat{Q}_h(s,a) - \hat{V}_h(s)\big)
\Big].
\end{equation}
This loss penalizes $\hat{V}_h(s) > \hat{Q}_h(s,a)$ more strongly than $\hat{V}_h(s) < \hat{Q}_h(s,a)$, so that
$\hat{V}_h(s)$ approximates a lower expectile of the hazard values of the dataset actions at
state $s$. Each hazard action-value network is updated by minimizing the squared residual
of the feasible Bellman operator,
\begin{equation}
\label{eq:qh_loss}
\mathcal{L}(\hat{Q}_{h,i})
=
\mathbb{E}_{(s,a,s')\sim\mathcal{D}}
\Big[
\big(
(1-\gamma)\,h(s) + \gamma \max\big(h(s), \hat{V}_h(s')\big) - \hat{Q}_{h,i}(s,a)
\big)^2
\Big],
\qquad i \in \{1,2\}.
\end{equation}

\subsection{Feasibility-Partitioned Conditional Actor Learning}
\label{sec:actor_learning}

\textbf{Behavior model.}
The actor is a CVAE with encoder $q_\phi(z \mid s,a)$ and decoder
$p_\theta(a \mid s,z) = \mathcal{N}\big(a;\, D_\theta(s,z),\, \sigma_{\mathrm{dec}}^2 I\big)$,
where $D_\theta(s,z)$ denotes the decoder mean.
For a fixed decoder variance, the reconstruction term of the ELBO in Eq.~(\ref{eq:elbo}) is
proportional to the squared error $\lVert a - D_\theta(s,z) \rVert_2^2$. For a dataset pair $(s,a)$, we write $\hat a := D_\theta(s,z)$ with
$z \sim q_\phi(\cdot \mid s,a)$ for the reconstructed action.

\textbf{Policy extraction.}
In safe offline RL, policy extraction is commonly done with
AWR \cite{peng2019advantage, koirala2024latent},
which weights the behavior-cloning likelihood of each dataset action by
$\exp\big(\hat A(s,a)/\tau\big)$ with temperature $\tau > 0$. Because the critic
error enters through the exponent, the variance of the AWR update
depends exponentially on the bias and the variance of the advantage error
(Theorem~\ref{theo:main_gradient_variance}). Practical implementations clip the exponential weights at a fixed maximum, which bounds the
variance but introduces bias, since actions whose weights reach the maximum are no longer
distinguished by their estimated advantages. FOCAL instead uses a first-order term that depends on the
critic through its action gradient, following the behavior-regularized policy gradient
of TD3+BC \cite{fujimoto2021minimalist}. Prior offline RL methods have combined CVAE behavior models with critic-gradient updates,
either by perturbing decoded actions \cite{fujimoto2019off} or by optimizing a policy in
the latent space \cite{zhou2021plas}. FOCAL differs in that the first-order term is
selected according to the estimated feasibility of each transition.

\textbf{Feasibility masks.}
Using the feasibility critics evaluated at the dataset transition $(s,a)$, we define the
feasibility and infeasibility masks
\begin{align}
\label{eq:masks}
\mathcal{M}_{\mathrm{feas}}(s,a)
&=
\mathbb{I}\big( \hat{V}_h(s) \le 0 \,\wedge\, \hat{Q}_h(s,a) \le 0 \big),
\\
\mathcal{M}_{\mathrm{infeas}}(s,a)
&=
\mathbb{I}\big( \hat{V}_h(s) > 0 \,\wedge\, \hat{Q}_h(s,a) > 0 \big),
\\
\mathcal{M}_{\mathrm{valid}}(s,a)
&=
\mathcal{M}_{\mathrm{feas}}(s,a) + \mathcal{M}_{\mathrm{infeas}}(s,a).
\end{align}
Two types of transitions are excluded. Transitions with $\hat{V}_h(s) > 0 \wedge \hat{Q}_h(s,a) \le 0$ are inconsistent with the
relation $V_h^*(s) = \min_{a} Q_h^*(s,a)$. Transitions with $\hat{V}_h(s) \le 0 \wedge \hat{Q}_h(s,a) > 0$ correspond to
actions estimated to be unsafe in states estimated to be feasible, and are not used for policy updates. 

\textbf{Objective.}
Combining behavior reconstruction, feasibility-dependent value optimization, and latent
regularization, the actor is trained by minimizing
\begin{equation}
\label{eq:cvae_objective}
\begin{split}
\mathcal{L}_{\mathrm{CVAE}}(\theta,\phi)
=
\mathbb{E}_{\substack{(s,a)\sim\mathcal{D}\\ z\sim q_{\phi}(\cdot \mid s,a)}}
\Big[
&\,\mathcal{M}_{\mathrm{feas}}(s,a)
\big( \lVert a-\hat{a} \rVert_2^2 - \lambda_r\, \hat{Q}_r(s,\hat{a}) \big)
\\
&+
\mathcal{M}_{\mathrm{infeas}}(s,a)
\big( \lVert a-\hat{a} \rVert_2^2 + \lambda_h\, \hat{Q}_h(s,\hat{a}) \big)
\\
&+
\mathcal{M}_{\mathrm{valid}}(s,a)\,
\beta\, D_{\mathrm{KL}}\big( q_{\phi}(z \mid s,a) \,\Vert\, \mathcal{N}(0,I) \big)
\Big],
\end{split}
\end{equation}
where $\hat a = D_\theta(s,z)$, $\beta > 0$ weights the latent regularization, and
\begin{equation}
\label{eq:step_sizes}
\lambda_r = \frac{\alpha_r}{\Phi_r},
\qquad
\lambda_h = \frac{\alpha_h}{\Phi_h},
\qquad
\Phi_r = \frac{1}{\lvert\mathcal{B}\rvert}\sum_{(s,a)\in\mathcal{B}} \big\lvert \hat{Q}_r(s,\hat a) \big\rvert,
\qquad
\Phi_h = \frac{1}{\lvert\mathcal{B}\rvert}\sum_{(s,a)\in\mathcal{B}} \big\lvert \hat{Q}_h(s,\hat a) \big\rvert,
\end{equation}
are normalized step sizes computed on each mini-batch $\mathcal{B}$, with $\alpha_r,
\alpha_h > 0$. As in TD3+BC \cite{fujimoto2021minimalist}, $\Phi_r$ and $\Phi_h$ are treated
as constants with respect to $(\theta,\phi)$. Since the objective involves no exponential weighting, FOCAL
does not require weight clipping, and the scale of its update is controlled by the
normalized step sizes in Eq.~(\ref{eq:step_sizes}).

The two branches of Eq.~(\ref{eq:cvae_objective}) assign distinct roles to the value terms: for transitions in the estimated feasible region, the reward term promotes higher estimated reward, while the reconstruction term constrains the reconstructed action to remain close to the dataset action. For transitions in the estimated infeasible region, the hazard term reduces estimated constraint violation, while the reconstruction term similarly limits deviation from the dataset action. The feasible branch excludes the hazard term because the dataset action already satisfies $\hat{Q}_h(s,a) \le 0$, and further reducing $\hat{Q}_h(s,\hat a)$ could shift the actor towards the interior of the estimated feasible region at the expense of task performance. Instead, the reconstruction term controls the hazard value of the training target by limiting its deviation from the dataset action, as established in Proposition~\ref{prop:cvae_dpg_stationary}.

\subsection{Adaptive Test-Time Inference}
\label{sec:adaptive_inference}
At test time, actions are generated by sampling latent variables from the prior and decoding
them. With a fixed prior scale, the latent variables are drawn from the same prior distribution in
every state. FOCAL instead sets the scale of the latent prior according to the estimated
feasibility of the state:
\setlength{\abovedisplayskip}{6pt}
\setlength{\belowdisplayskip}{6pt}
\begin{equation}
\label{eq:adaptive_variance}
\sigma_z(s) = \mathrm{clip}\big(-\hat{V}_h(s),\, \sigma_{\min},\, 1\big),
\end{equation}
where $\sigma_{\min} \in (0,1)$. Since $h(s) \ge -1$ under the definition in
Eq.~(\ref{eq:h_definition}), the exact value function satisfies $V_h^*(s) \ge -1$, with
equality when no violation is reachable. For $\hat{V}_h(s) \in [-1, -\sigma_{\min}]$, the
scale satisfies $\sigma_z(s) = -\hat{V}_h(s)$. States with no estimated reachable violation
($\hat{V}_h(s) \le -1$) are therefore assigned the standard prior scale $\sigma_z(s) = 1$, and
the scale decreases linearly as $\hat{V}_h(s)$ increases towards the feasibility boundary
$\hat{V}_h(s) = 0$. States with $\hat{V}_h(s) \ge -\sigma_{\min}$, which include all states
estimated to be infeasible, are assigned the minimum scale $\sigma_{\min}$.
A smaller scale concentrates the latent samples around the mode of the prior.

Given state $s$, we draw $N$ latent samples $z_i \sim \mathcal{N}\big(0, \sigma_z(s)^2 I\big)$
and decode them into candidate actions $\hat a_i = D_\theta(s, z_i)$, $i = 1,\dots,N$. Let
$\mathcal{C}(s) = \{\, i : \hat{Q}_h(s,\hat a_i) \le 0 \,\}$ denote the indices of candidates
estimated to be feasible. The executed action is $a^{\mathrm{exec}} = \hat a_{i^\star}$, where
\begin{equation}
\label{eq:action_selection}
i^\star =
\begin{cases}
\displaystyle \arg\max_{i \in \mathcal{C}(s)} \hat{Q}_r(s,\hat a_i), & \text{if } \mathcal{C}(s) \neq \emptyset,\\[8pt]
\displaystyle \arg\min_{i \in \{1,\dots,N\}} \hat{Q}_h(s,\hat a_i), & \text{if } \mathcal{C}(s) = \emptyset.
\end{cases}
\end{equation}

Candidate filtering returns a feasible action only if one is among the candidates and the
hazard critic identifies it. If $\hat{Q}_h = Q_h^*$ and the sampling distribution assigns
positive probability to feasible actions, a feasible candidate is obtained with
probability approaching one as $N \rightarrow \infty$. Since the critics are learned from
finite data and $N$ is finite, filtering reduces, but does not eliminate, constraint
violations.

% ---------------------------------------------------------------------------
\setlength{\abovedisplayskip}{4pt}
\setlength{\belowdisplayskip}{4pt}
\subsection{Theoretical Analysis}
\label{sec:theory}

\subsubsection{Action Deviation}

\begin{assumption}[Lipschitz continuous critics]
\label{assum:critic_lipschitz}
The learned critics $\hat{Q}_r(s,\cdot)$ and $\hat{Q}_h(s,\cdot)$ are $L_{Q_r}$- and
$L_{Q_h}$-Lipschitz continuous in the action, so that
$\lVert \nabla_a \hat{Q}_r(s,a) \rVert_2 \le L_{Q_r}$ and
$\lVert \nabla_a \hat{Q}_h(s,a) \rVert_2 \le L_{Q_h}$ almost everywhere.
\end{assumption}

\begin{proposition}[Bounded action deviation and hazard value of the training target]
\label{prop:cvae_dpg_stationary}
Let Assumption~\ref{assum:critic_lipschitz} hold, let $(s,a)$ be a transition with
$\mathcal{M}_{\mathrm{valid}}(s,a) = 1$, and treat $\lambda_r$ and $\lambda_h$ as constants.
Let $\hat a^*$ be a stationary point, in the interior of the action space, of the
region-specific loss $\ell_{\mathrm{feas}}(\hat a) = \lVert a - \hat a \rVert_2^2 - \lambda_r
\hat{Q}_r(s,\hat a)$ if $\mathcal{M}_{\mathrm{feas}}(s,a) = 1$, or
$\ell_{\mathrm{infeas}}(\hat a) = \lVert a - \hat a \rVert_2^2 + \lambda_h \hat{Q}_h(s,\hat a)$ if
$\mathcal{M}_{\mathrm{infeas}}(s,a) = 1$. Then
\begin{align}
\text{(Feasible)} \quad
&\hat a^* = a + \frac{\lambda_r}{2}\, \nabla_{\hat a} \hat{Q}_r(s,\hat a^*),
\qquad
\lVert \hat a^* - a \rVert_2 \le \frac{\lambda_r L_{Q_r}}{2},
\label{eq:prop2_feas}
\\
&\hat{Q}_h(s,\hat a^*) \le \hat{Q}_h(s,a) + \frac{\lambda_r L_{Q_r} L_{Q_h}}{2} \le \frac{\lambda_r L_{Q_r} L_{Q_h}}{2},
\label{eq:prop2_hazard}
\\
\text{(Infeasible)} \quad
&\hat a^* = a - \frac{\lambda_h}{2}\, \nabla_{\hat a} \hat{Q}_h(s,\hat a^*),
\qquad
\lVert \hat a^* - a \rVert_2 \le \frac{\lambda_h L_{Q_h}}{2},
\label{eq:prop2_infeas}
\end{align}
The bounds characterize the training target defined by the region-specific loss. They do
not apply to actions decoded at test time.
\end{proposition}

\textit{Proof.} See Appendix~\ref{subsubsec:stationary_point}.

\subsubsection{Dependence of the Policy Update on Critic Estimation Error}

\begin{assumption}[Critic estimation error]
\label{assum:critic_error}
Fix a state $s$. The learned critic, denoted by $\hat Q \in \{\hat Q_r, \hat Q_h\}$, is modeled as a random function of the action,
$\hat Q(s,\cdot) = Q^*(s,\cdot) + \epsilon_Q(s,\cdot)$, where the randomness (denoted by $\epsilon$) captures variations across critic training runs (e.g., initialization and batch sampling). For any action $a'$, the error has mean bias $\mu(s,a') := \mathbb{E}_{\epsilon}[\epsilon_Q(s,a')]$ and finite variance $\sigma^2(s,a') := \mathrm{Var}_{\epsilon}[\epsilon_Q(s,a')]$. We denote by $\xi(s,a') := \epsilon_Q(s,a') - \mu(s,a')$ the zero-mean fluctuation.
\end{assumption}

\begin{assumption}[Advantage error for AWR]
\label{assum:adv_error}
For a dataset transition $(s,a)$, the advantage estimation error $\epsilon_A(s,a) := \hat A(s,a) - A^*(s,a)$ satisfies $\epsilon_A(s,a) \sim \mathcal{N}\big(\mu_A(s,a), \sigma_A^2(s,a)\big)$.
\end{assumption}

\begin{assumption}[Gaussian decoder and actor geometry]
\label{assum:decoder}
For a dataset pair $(s,a)$, let $\tilde\pi_\psi(\cdot \mid s,a)$, with $\psi = (\theta,\phi)$, denote the distribution of $D_\theta(s,z) + \sigma_{\mathrm{dec}}\,\eta$, where $z \sim q_\phi(\cdot \mid s,a)$, $\eta \sim \mathcal{N}(0, I_d)$, $d = \dim(\mathcal{A})$, and $\sigma_{\mathrm{dec}} > 0$ is a fixed standard deviation. The decoder Jacobian $J_\theta := \nabla_\theta D_\theta(s,z)$ satisfies $\lVert J_\theta \rVert_2 \le L_\theta$, and differentiation with respect to $\psi$ can be interchanged with integration over $\tilde\pi_\psi$. We denote by $u(\hat a) := \nabla_\psi \log \tilde\pi_\psi(\hat a \mid s,a)$ the actor score function, by $F(\psi;s,a) := \mathbb{E}_{\hat a \sim \tilde\pi_\psi}\big[ u(\hat a)\, u(\hat a)^{\top} \big]$ the Fisher information matrix of the actor, and by $F_q(\phi;s,a)$ the Fisher information of the encoder $q_\phi(\cdot \mid s,a)$ with respect to $\phi$.
\end{assumption}

\begin{theorem}[Sensitivity of Policy Updates to Critic Error]
\label{theo:main_gradient_variance}
Fix a dataset transition $(s,a)$, and let $\mathbb{E}_\epsilon$ denote expectation over critic training randomness.

\textbf{(i) AWR.} We analyze AWR with the unclipped exponential weight. Let $\tau > 0$ be the temperature parameter, and let $v := \nabla_\psi \log \pi_\psi(a \mid s)$. Define $g_{\mathrm{AWR}} := \exp\big(\hat A(s,a)/\tau\big)\, v$ and $g^*_{\mathrm{AWR}} := \exp\big(A^*(s,a)/\tau\big)\, v$. Under Assumption~\ref{assum:adv_error}, with $A^* = A^*(s,a)$, $\mu_A = \mu_A(s,a)$, and $\sigma_A = \sigma_A(s,a)$:
\begin{align}
\big\lVert \mathbb{E}_\epsilon[g_{\mathrm{AWR}}] - g^*_{\mathrm{AWR}} \big\rVert_2
&= \lVert v \rVert_2\, e^{A^*/\tau}\, \Big\lvert e^{\mu_A/\tau + \sigma_A^2/(2\tau^2)} - 1 \Big\rvert, \label{eq:awr_bias} \\
\mathbb{E}_\epsilon\Big[ \big\lVert g_{\mathrm{AWR}} - \mathbb{E}_\epsilon[g_{\mathrm{AWR}}] \big\rVert_2^2 \Big]
&= \lVert v \rVert_2^2\, e^{2A^*/\tau}\, e^{2\mu_A/\tau + \sigma_A^2/\tau^2} \Big( e^{\sigma_A^2/\tau^2} - 1 \Big). \label{eq:awr_var}
\end{align}

\textbf{(ii) FOCAL.}  Let $g_{\mathrm{FOCAL}} := \lambda\, \nabla_\psi\, \mathbb{E}_{\hat a \sim \tilde\pi_\psi(\cdot \mid s,a)}\big[ \hat Q(s,\hat a) \big]$ with error-free counterpart $g^*_{\mathrm{FOCAL}} := \lambda\, \nabla_\psi\, \mathbb{E}_{\hat a \sim \tilde\pi_\psi(\cdot \mid s,a)}\big[ Q^*(s,\hat a) \big]$. Under Assumptions~\ref{assum:critic_error} and~\ref{assum:decoder}:
\begin{align}
\big\lVert \mathbb{E}_\epsilon[g_{\mathrm{FOCAL}}] - g^*_{\mathrm{FOCAL}} \big\rVert_2
&\le \lambda\, \sqrt{\operatorname{tr} F(\psi;s,a)}\, \sqrt{\mathrm{Var}_{\hat a \sim \tilde\pi_\psi}\big[ \mu(s,\hat a) \big]}, \label{eq:focal_bias} \\
\mathbb{E}_\epsilon\Big[ \big\lVert g_{\mathrm{FOCAL}} - \mathbb{E}_\epsilon[g_{\mathrm{FOCAL}}] \big\rVert_2^2 \Big]
&\le \lambda^2\, \operatorname{tr} F(\psi;s,a)\, \mathbb{E}_\epsilon\Big[ \mathrm{Var}_{\hat a \sim \tilde\pi_\psi}\big[ \xi(s,\hat a) \big] \Big] \nonumber \\
&\le \lambda^2\, \operatorname{tr} F(\psi;s,a)\, \mathbb{E}_{\hat a \sim \tilde\pi_\psi}\big[ \sigma^2(s,\hat a) \big], \label{eq:focal_var}
\end{align}
where the trace of the Fisher information is bounded by a decoder term and the Fisher
information of the encoder:
\begin{equation}
\operatorname{tr} F(\psi;s,a) \le \frac{d\, L_\theta^2}{\sigma_{\mathrm{dec}}^2}
+ \operatorname{tr} F_q(\phi;s,a). \label{eq:fisher}
\end{equation}
The same bounds hold for the infeasible branch with $(\hat Q, \lambda)$ replaced by $(\hat Q_h, \lambda_h)$.
\end{theorem}

\textit{Proof.} See Appendix~\ref{subsubsec:variance_theorem}.
\section{Results and Discussion}
\label{sec:results}

\begin{table*}[t!]
    \centering
    \caption{Normalized performance on the DSRL benchmark \cite{liu2023datasets}, averaged across 3 seeds (20 episodes each). $\pm$ represents standard deviation across 3 training runs. Arrows ($\uparrow$/$\downarrow$) indicate the preferred direction for each metric. \textcolor{gray}{Gray}: Unsafe policies. \textbf{Bold}: Safe policies (normalized cost $\le 1.0$). \textbf{\textcolor{blue}{Blue}}: Safe policies achieving the highest reward.}
    \label{tab:main_results}
    \resizebox{\textwidth}{!}{
    \begin{tabular}{l *{16}{c}}
        \toprule
        \multirow{2}{*}{Task} & \multicolumn{2}{c}{BC} & \multicolumn{2}{c}{CDT} & \multicolumn{2}{c}{BCQ-Lag} & \multicolumn{2}{c}{CPQ} & \multicolumn{2}{c}{COptiDICE} & \multicolumn{2}{c}{LSPC} & \multicolumn{2}{c}{FISOR} &  \multicolumn{2}{c}{FOCAL (ours)} \\
        \cmidrule(lr){2-3} \cmidrule(lr){4-5} \cmidrule(lr){6-7} \cmidrule(lr){8-9} \cmidrule(lr){10-11} \cmidrule(lr){12-13} \cmidrule(lr){14-15} \cmidrule(lr){16-17}
        & reward $\uparrow$ & cost $\downarrow$ & reward $\uparrow$ & cost $\downarrow$ & reward $\uparrow$ & cost $\downarrow$ & reward $\uparrow$ & cost $\downarrow$ & reward $\uparrow$ & cost $\downarrow$ & reward $\uparrow$ & cost $\downarrow$ & reward $\uparrow$ & cost $\downarrow$ & reward $\uparrow$ & cost $\downarrow$ \\
        \midrule
        CarButton1     & \gy{0.01\sd{0.03}}  & \gy{6.19\sd{0.39}}  & \gy{0.17\sd{0.02}}  & \gy{7.05\sd{0.12}}  & \gy{0.13\sd{0.05}}  & \gy{6.68\sd{0.59}}  & \gy{0.22\sd{0.05}}  & \gy{40.06\sd{5.71}} & \gy{-0.16\sd{0.09}} & \gy{4.63\sd{1.29}}  & \gy{-0.04\sd{0.23}}  & \gy{3.36\sd{1.42}}  & \bd{-0.02\sd{0.05}} & \bd{0.26\sd{0.12}} & \bb{-0.01\sd{0.03}} & \bb{0.72\sd{0.14}}   \\
        CarButton2     & \gy{-0.10\sd{0.02}} & \gy{4.47\sd{0.30}}  & \gy{0.23\sd{0.01}}  & \gy{12.87\sd{0.02}} & \gy{-0.04\sd{0.05}} & \gy{4.43\sd{1.19}}  & \gy{0.08\sd{0.11}}  & \gy{19.03\sd{8.54}} & \gy{-0.17\sd{0.06}} & \gy{3.40\sd{1.10}}  & \gy{-0.11\sd{0.28}}  & \gy{2.09\sd{2.06}} & \bb{0.01\sd{0.01}}  & \bb{0.58\sd{0.34}} & \bd{0.00\sd{0.02}} & \bd{0.91\sd{0.41}}   \\
        CarGoal1       & \gy{0.35\sd{0.08}}  & \gy{1.54\sd{0.11}}  & \gy{0.60\sd{0.01}}  & \gy{3.15\sd{0.17}}  & \gy{0.44\sd{0.05}}  & \gy{2.76\sd{0.50}}  & \gy{0.33\sd{0.07}}  & \gy{4.93\sd{0.81}}  & \gy{0.43\sd{0.06}}  & \gy{2.81\sd{0.33}}  &  \gy{0.28\sd{0.26}} & \gy{1.17\sd{0.44}}  & \bd{0.49\sd{0.03}}  & \bd{0.83\sd{0.57}} & \bb{0.61\sd{0.06}} & \bb{0.80\sd{0.33}}   \\
        CarGoal2       & \gy{0.22\sd{0.05}}  & \gy{3.30\sd{0.26}}  & \gy{0.45\sd{0.01}}  & \gy{6.05\sd{0.14}}  & \gy{0.34\sd{0.05}}  & \gy{4.72\sd{0.99}}  & \gy{0.10\sd{0.20}}  & \gy{6.31\sd{2.00}}  & \gy{0.19\sd{0.04}}  & \gy{2.83\sd{0.41}}  & \gy{0.17}\sd{0.21}  & \gy{2.16\sd{0.77}}  & \bd{0.06\sd{0.03}}  & \bd{0.33\sd{0.33}} & \bb{0.21\sd{0.03}} & \bb{0.65\sd{1.10}}   \\
        AntVel         & \gy{0.99\sd{0.01}}  & \gy{12.19\sd{0.10}} & \bb{0.98\sd{0.00}}  & \bb{0.91\sd{0.12}}  & \gy{0.85\sd{0.01}}  & \gy{18.54\sd{1.63}} & \bd{-1.01\sd{0.00}} & \bd{0.00\sd{0.00}}  & \gy{1.00\sd{0.00}}  & \gy{10.29\sd{2.01}} &  \bb{0.98\sd{0.05}} & \bb{0.69\sd{0.15}}  & \bd{0.89\sd{0.04}}  & \bd{0.00\sd{0.00}} & \bd{0.81\sd{0.01}} & \bd{0.00\sd{0.00}}   \\
        HalfCheetahVel & \gy{0.97\sd{0.03}}  & \gy{17.93\sd{0.63}} & \bd{0.97\sd{0.01}}  & \bd{0.55\sd{0.01}}  & \gy{1.04\sd{0.07}}  & \gy{57.06\sd{8.29}} & \gy{0.08\sd{0.14}}  & \gy{2.56\sd{0.19}}  & \bd{0.43\sd{0.01}}  & \bd{0.00\sd{0.00}}  & \bb{0.98\sd{0.02}}  & \bb{0.31\sd{0.54}}  & \bd{0.89\sd{0.02}}  & \bd{0.00\sd{0.00}} & \bd{0.95\sd{0.00}} & \bd{0.00\sd{0.00}}   \\
        SwimmerVel     & \gy{0.38\sd{0.20}}  & \gy{2.98\sd{0.07}}  & \gy{0.67\sd{0.01}}  & \gy{1.47\sd{0.08}}  & \gy{0.29\sd{0.33}}  & \gy{4.10\sd{0.95}}  & \gy{0.31\sd{0.06}}  & \gy{11.58\sd{0.96}} & \gy{0.58\sd{0.06}}  & \gy{23.64\sd{1.77}} & \bb{0.62\sd{0.21}}  & \bb{0.62\sd{1.07}}  & \bd{-0.04\sd{0.01}} & \bd{0.00\sd{0.00}} & \bd{0.13\sd{0.05}} & \bd{0.00\sd{0.00}}  \\
        HopperVel      & \gy{0.78\sd{0.13}}  & \gy{14.27\sd{0.27}}  & \bd{0.64\sd{0.06}} & \bd{0.72\sd{0.08}}  &  \gy{0.81\sd{0.09}} & \gy{14.96\sd{3.43}} & \gy{0.25\sd{0.09}} & \gy{15.02\sd{2.29}} & \gy{0.42\sd{0.06}} & \gy{10.97\sd{1.54}} & \bd{0.73\sd{0.04}} & \bd{0.45\sd{0.78}}  & \bd{0.14\sd{0.02}} & \bd{0.55\sd{0.23}} & \bb{0.83\sd{0.04}} & \bb{0.00\sd{0.00}}   \\
        Walker2dVel    & \gy{0.78\sd{0.05}}  &  \gy{6.19\sd{0.32}} & \bd{0.75\sd{0.09}} & \bd{0.12\sd{0.21}}  & \bb{0.80\sd{0.01}} & \bb{0.00\sd{0.00}} & \gy{0.06\sd{0.05}} & \gy{4.10\sd{0.09}} & \gy{0.14\sd{0.01}} & \gy{4.04\sd{0.07}} & \bb{0.80\sd{0.01}} & \bb{0.76\sd{0.24}} & \bd{0.57\sd{0.12}} & \bd{0.97\sd{0.59}} & \bd{0.72\sd{0.08}} & \bd{0.42\sd{0.22}}   \\
        \midrule
        \makecell{\bd{SafetyGym} \\ \bd{Average}} & \gy{0.49} & \gy{7.67} & \gy{0.61} & \gy{3.65} & \gy{0.52} & \gy{12.58} & \gy{0.05} & \gy{11.51} & \gy{0.32} & \gy{6.96} & \gy{0.49} & \gy{1.29} & \bd{0.33} & \bd{0.39} & \bb{0.47} & \bb{0.39} \\
        \midrule
        AntRun         & \gy{0.73\sd{0.15}}  & \gy{11.73\sd{0.84}} & \gy{0.70\sd{0.04}}  & \gy{1.88\sd{0.42}}  & \gy{0.65\sd{0.07}}  & \gy{3.30\sd{2.39}}  & \bd{0.00\sd{0.02}}  & \bd{0.00\sd{0.00}}  & \gy{0.62\sd{0.01}}  & \gy{3.64\sd{0.69}}  &  \bd{0.15\sd{0.13}} & \bd{0.00\sd{0.00}}  & \bb{0.45\sd{0.03}}  & \bb{0.03\sd{0.00}} & \bd{0.34\sd{0.04}} & \bd{0.00\sd{0.00}}   \\
        BallRun        & \gy{0.67\sd{0.14}}  & \gy{11.38\sd{0.39}} & \bb{0.32\sd{0.09}}  & \bb{0.45\sd{0.19}}  & \gy{0.43\sd{0.01}}  & \gy{6.25\sd{0.35}}  & \gy{0.85\sd{0.00}}  & \gy{13.67\sd{0.12}} & \gy{0.55\sd{0.00}}  & \gy{11.32\sd{0.00}} &  \bd{0.16\sd{0.01}} & \bd{0.00\sd{0.00}}  & \bd{0.18\sd{0.07}}  & \bd{0.00\sd{0.00}} & \bd{0.28\sd{0.03}} & \bd{0.00\sd{0.00}}    \\
        CarRun         & \gy{0.96\sd{0.00}}  & \gy{1.88\sd{0.02}}  & \gy{0.99\sd{0.01}}  & \gy{1.10\sd{0.31}}  & \gy{0.84\sd{0.01}}  & \gy{2.51\sd{0.91}}  & \gy{1.06\sd{0.01}}  & \gy{10.49\sd{0.18}} & \bb{0.92\sd{0.00}}  & \bb{0.00\sd{0.00}}  &  \gy{0.97\sd{0.01}} &  \gy{2.06\sd{1.04}} & \bd{0.73\sd{0.03}}  & \bd{0.14\sd{0.01}} & \bd{0.83\sd{0.01}} & \bd{0.00\sd{0.00}}    \\
        DroneRun       & \gy{0.55\sd{0.25}}  & \gy{5.21\sd{0.97}}  & \bb{0.58\sd{0.04}}  & \bb{0.30\sd{0.68}}  & \gy{0.80\sd{0.12}}  & \gy{17.98\sd{0.81}} & \gy{0.02\sd{0.10}}  & \gy{7.95\sd{0.58}}  & \gy{0.72\sd{0.02}}  & \gy{13.77\sd{0.10}} & \bd{0.56\sd{0.06}}  & \bd{0.29\sd{0.51}}  & \bd{0.30\sd{0.12}}  & \bd{0.55\sd{0.07}} & \bb{0.58\sd{0.02}} & \bb{0.58\sd{0.38}}    \\
        AntCircle      & \gy{0.65\sd{0.16}}  & \gy{19.45\sd{2.67}} & \gy{0.48\sd{0.20}}  & \gy{7.44\sd{4.33}}  & \gy{0.67\sd{0.25}}  & \gy{19.13\sd{3.08}} & \bd{0.00\sd{0.00}}  & \bd{0.00\sd{0.00}}  & \gy{0.18\sd{0.10}}  & \gy{13.41\sd{6.70}} & \bd{0.24\sd{0.14}}  &  \bd{0.17\sd{0.30}} & \bd{0.20\sd{0.05}}  & \bd{0.00\sd{0.00}} & \bb{0.52\sd{0.06}} & \bb{0.76\sd{0.15}}   \\
        BallCircle     & \gy{0.72\sd{0.08}}  & \gy{10.02\sd{0.17}} & \gy{0.68\sd{0.06}}  & \gy{2.10\sd{0.27}}  & \gy{0.67\sd{0.11}}  & \gy{8.50\sd{1.04}}  & \gy{0.40\sd{0.01}}  & \gy{4.37\sd{0.00}}  & \gy{0.70\sd{0.04}}  & \gy{9.06\sd{0.79}} & \bd{0.41\sd{0.28}}  & \bd{0.00\sd{0.00}}  & \bd{0.34\sd{0.12}}  & \bd{0.00\sd{0.00}} & \bb{0.55\sd{0.00}} & \bb{0.00\sd{0.00}}  \\
        CarCircle      & \gy{0.65\sd{0.22}}  & \gy{11.16\sd{0.67}} & \gy{0.71\sd{0.06}}  & \gy{2.19\sd{0.61}}  & \gy{0.68\sd{0.19}}  & \gy{8.84\sd{1.37}}  & \gy{0.49\sd{0.02}}  & \gy{4.48\sd{0.00}}  & \gy{0.44\sd{0.05}}  & \gy{7.73\sd{2.98}}  & \bd{0.62\sd{0.12}}  & \bd{0.05\sd{0.09}}  & \bd{0.40\sd{0.10}}  & \bd{0.11\sd{0.01}} & \bb{0.71\sd{0.03}} & \bb{0.00\sd{0.00}}   \\
        DroneCircle    & \gy{0.82\sd{0.18}}  & \gy{13.78\sd{0.27}} & \gy{0.55\sd{0.07}}  & \gy{1.29\sd{0.27}}  & \gy{0.95\sd{0.07}}  & \gy{18.56\sd{0.89}} & \gy{-0.27\sd{0.05}} & \gy{1.29\sd{0.97}}  & \gy{0.24\sd{0.03}}  & \gy{2.19\sd{0.46}}  & \bd{0.34\sd{0.02}}  & \bd{0.00\sd{0.00}}  & \bd{0.48\sd{0.04}}  & \bd{0.00\sd{0.00}} & \bb{0.54\sd{0.02}} & \bb{0.00\sd{0.00}}   \\
        \midrule
        \makecell{\bd{BulletGym} \\ \bd{Average}} & \gy{0.72} & \gy{10.58} & \gy{0.63} & \gy{2.09} & \gy{0.71} & \gy{10.63} & \gy{0.32} & \gy{5.28} & \gy{0.55} & \gy{7.64} & \bd{0.43} & \bd{0.32} & \bd{0.39} & \bd{0.10} & \bb{0.54} & \bb{0.17} \\
        \midrule
        EasySparse     & \gy{0.32\sd{0.08}}  & \gy{4.73\sd{0.02}}  & \bd{0.05\sd{0.14}}  & \bd{0.10\sd{0.32}}  & \gy{0.99\sd{0.00}}  & \gy{14.00\sd{0.06}} & \bd{-0.06\sd{0.00}} & \bd{0.24\sd{0.02}}  & \gy{0.94\sd{0.02}}  & \gy{18.21\sd{0.27}} &  \gy{0.50\sd{0.14}} & \gy{1.57\sd{1.43}}  & \bb{0.38\sd{0.02}}  & \bb{0.53\sd{0.05}} &  \bd{0.26\sd{0.03}} & \bd{0.31\sd{0.09}} \\
        EasyMean       & \gy{0.22\sd{0.03}}  & \gy{2.68\sd{0.02}}  & \bd{0.27\sd{0.11}}  & \bd{0.24\sd{0.55}}  & \gy{0.54\sd{0.06}}  & \gy{10.35\sd{0.35}} & \bd{-0.06\sd{0.00}} & \bd{0.24\sd{0.01}}  & \gy{0.74\sd{0.16}}  & \gy{14.81\sd{1.47}} & \gy{0.58\sd{0.06}}  & \gy{1.45\sd{0.86}}  & \bb{0.38\sd{0.01}}  & \bb{0.25\sd{0.21}} & \bd{0.26\sd{0.07}} & \bd{0.51\sd{0.52}} \\
        EasyDense      & \gy{0.20\sd{0.07}}  & \gy{1.70\sd{0.01}}  & \gy{0.43\sd{0.18}}  & \gy{2.31\sd{0.43}}  & \gy{0.40\sd{0.00}}  & \gy{6.64\sd{0.01}}  & \bd{-0.06\sd{0.00}} & \bd{0.29\sd{0.01}}  & \gy{0.60\sd{0.10}}  & \gy{11.27\sd{0.53}} & \gy{0.57\sd{0.10}}  &  \gy{1.81\sd{1.20}} & \bb{0.36\sd{0.04}}  & \bb{0.25\sd{0.15}} & \bd{0.30\sd{0.02}} & \bd{0.54\sd{0.31}} \\
        MediumSparse   & \gy{0.53\sd{0.34}}  & \gy{1.74\sd{0.32}}  & \gy{0.26\sd{0.11}}  & \gy{2.20\sd{0.26}}  & \gy{0.93\sd{0.00}}  & \gy{7.48\sd{0.02}}  & \bd{-0.08\sd{0.02}} & \bd{0.18\sd{0.03}}  & \gy{0.64\sd{0.37}}  & \gy{7.26\sd{1.90}}  & \bb{0.94\sd{0.02}}  & \bb{0.91\sd{0.68}} & \bd{0.42\sd{0.09}}  & \bd{0.22\sd{0.07}} & \bd{0.64\sd{0.34}}  & \bd{0.92\sd{0.11}} \\
        MediumMean     & \gy{0.66\sd{0.06}}  & \gy{2.94\sd{0.00}}  & \gy{0.28\sd{0.39}}  & \gy{2.13\sd{0.83}}  & \gy{0.60\sd{0.12}}  & \gy{6.35\sd{0.21}}  & \bd{-0.08\sd{0.00}} & \bd{0.28\sd{0.02}}  & \gy{0.73\sd{0.34}}  & \gy{8.35\sd{0.92}}  & \gy{0.90\sd{0.12}}  & \gy{1.21\sd{0.51}}  & \bd{0.39\sd{0.20}}  & \bd{0.08\sd{0.12}} & \bb{0.65\sd{0.08}} & \bb{0.94\sd{0.04}} \\
        MediumDense    & \gy{0.65\sd{0.00}}  & \gy{3.79\sd{0.00}}  & \bd{0.29\sd{0.12}}  & \bd{0.77\sd{0.71}}  & \gy{0.64\sd{0.21}}  & \gy{3.78\sd{1.19}}  & \bd{-0.08\sd{0.00}} & \bd{0.20\sd{0.01}}  & \gy{0.91\sd{0.13}}  & \gy{9.52\sd{0.65}}  & \bb{0.81\sd{0.30}}  &  \bb{0.68\sd{0.71}} & \bd{0.49\sd{0.06}}  & \bd{0.44\sd{0.13}} & \bd{0.42\sd{0.13}} & \bd{0.79\sd{0.13}} \\
        HardSparse     & \gy{0.28\sd{0.05}}  & \gy{1.98\sd{0.10}}  & \bd{0.17\sd{0.08}}  & \bd{0.47\sd{0.33}}  & \gy{0.48\sd{0.04}}  & \gy{7.52\sd{0.05}}  & \bd{-0.04\sd{0.00}} & \bd{0.28\sd{0.01}}  & \gy{0.34\sd{0.10}}  & \gy{7.34\sd{0.27}}  & \gy{0.19\sd{0.09}}  & \gy{1.20\sd{0.51}}  & \bb{0.30\sd{0.02}}  & \bb{0.01\sd{0.00}} & \bd{0.24\sd{0.20}}  & \bd{0.89\sd{0.22}} \\
        HardMean       & \gy{0.34\sd{0.00}}  & \gy{3.76\sd{0.00}}  & \gy{0.28\sd{0.21}}  & \gy{3.32\sd{0.31}}  & \gy{0.31\sd{0.13}}  & \gy{6.06\sd{0.72}}  & \bd{-0.05\sd{0.00}} & \bd{0.24\sd{0.02}}  & \gy{0.36\sd{0.19}}  & \gy{7.51\sd{2.00}}  & \bb{0.33\sd{0.14}}  & \bb{0.84\sd{1.19}}  & \bd{0.26\sd{0.04}}  & \bd{0.09\sd{0.05}} & \bd{0.20\sd{0.06}}  & \bd{0.86\sd{0.28}} \\
        HardDense      & \gy{0.40\sd{0.06}}  & \gy{5.57\sd{0.01}}  & \gy{0.24\sd{0.15}}  & \gy{1.49\sd{0.42}}  & \gy{0.39\sd{0.03}}  & \gy{5.11\sd{0.14}}  & \bd{-0.04\sd{0.01}} & \bd{0.24\sd{0.01}}  & \gy{0.42\sd{0.21}}  & \gy{8.11\sd{2.15}}  &  \gy{0.52\sd{0.19}} & \gy{1.28\sd{0.93}}  & \bb{0.30\sd{0.02}}  & \bb{0.34\sd{0.18}} &  \bb{0.30\sd{0.09}} & \bb{0.59\sd{0.44}} \\
        \midrule
        \makecell{\bd{MetaDrive} \\ \bd{Average}} & \gy{0.40} & \gy{3.21} & \gy{0.25} & \gy{1.45} & \gy{0.59} & \gy{7.48} & \bd{-0.06} & \bd{0.24} & \gy{0.63} & \gy{10.26} & \gy{0.59} & \gy{1.22} & \bb{0.36} & \bb{0.25} & \bd{0.35} & \bd{0.70} \\
        \bottomrule
    \end{tabular}
    }
\end{table*}

% Requires: graphicx, subcaption, booktabs, multirow, xcolor.
% Placeholders that must be filled before submission are marked with \todo{...}.
\providecommand{\todo}[1]{\textcolor{red}{[TODO: #1]}}

\subsection{Experimental Setup}
\label{sec:setup}

\textbf{Benchmark and evaluation protocol.}
We evaluate FOCAL on the Datasets for Safe Reinforcement Learning (DSRL)
benchmark~\cite{liu2023datasets}, using tasks from Safety-Gymnasium~\cite{ji2023safety},
Bullet-Safety-Gym~\cite{gronauer2022bullet}, and MetaDrive~\cite{li2022metadrive}.
Although FOCAL uses state-wise feasibility information during training, we follow the
standard DSRL evaluation protocol to enable consistent comparison with existing baselines.
Performance is measured by the normalized return
$R = (R_\pi - R_{\min}) / (R_{\max} - R_{\min})$, where $R_\pi$ is the undiscounted
episodic return and $R_{\max}$ and $R_{\min}$ are the maximum and minimum episodic returns
in the dataset, and by the normalized cost $C = C_\pi / \kappa$, where $C_\pi$ is the
undiscounted episodic cost and $\kappa > 0$ is the cost limit. Following prior
work~\cite{liu2023datasets, zheng2024safe}, a policy is considered safe if its normalized
cost, averaged over evaluation episodes, satisfies $C \le 1$. We use the cost limits of
$\kappa = 10$ for Safety-Gymnasium and $\kappa = 5$ for
Bullet-Safety-Gym and MetaDrive \cite{zheng2024safe}.

Since the DSRL criterion constrains the average episodic cost, the results below report
empirical safety under this protocol for all methods, rather than state-wise constraint
satisfaction. All results are averaged over three independent training runs, each evaluated
on $20$ episodes.

\textbf{Baselines.}
We compare FOCAL with behavior cloning (BC), which imitates the entire dataset; the
Constrained Decision Transformer (CDT)~\cite{liu2023constrained}, which performs sequence
modeling conditioned on target returns and costs; BCQ-Lag and CPQ~\cite{xu2022constraints},
which incorporate cost penalties into value-based learning; COptiDICE~\cite{lee2022coptidice},
which enforces constraints through stationary distribution correction; LSPC~\cite{koirala2024latent},
a CVAE-based policy that uses AWR for policy extraction; and FISOR~\cite{zheng2024safe},
which combines HJ reachability with a diffusion policy extracted by weighted regression.
Implementation details are given in Appendix~\ref{app:experimental_details}.

\subsection{Main Results}
\label{sec:main_results}

Table~\ref{tab:main_results} reports normalized return and cost on the DSRL tasks. FOCAL satisfies the prescribed threshold-based safety constraints across all evaluated environments. Among the compared methods, only FISOR and FOCAL consistently adhere to the safety requirements, while other baselines frequently violate the cost limits. In terms of reward, FOCAL achieves competitive or higher normalized returns compared to baselines across most tasks. These results indicate that HJ reachability with a behavior-regularized first-order objective within a generative policy can effectively maintain constraint satisfaction while achieving strong task performance.

\subsection{Ablation Studies}
\label{sec:ablations}

\begin{figure}[t]
    \centering
    \begin{subfigure}[b]{0.48\textwidth}
        \centering
        \includegraphics[width=\textwidth]{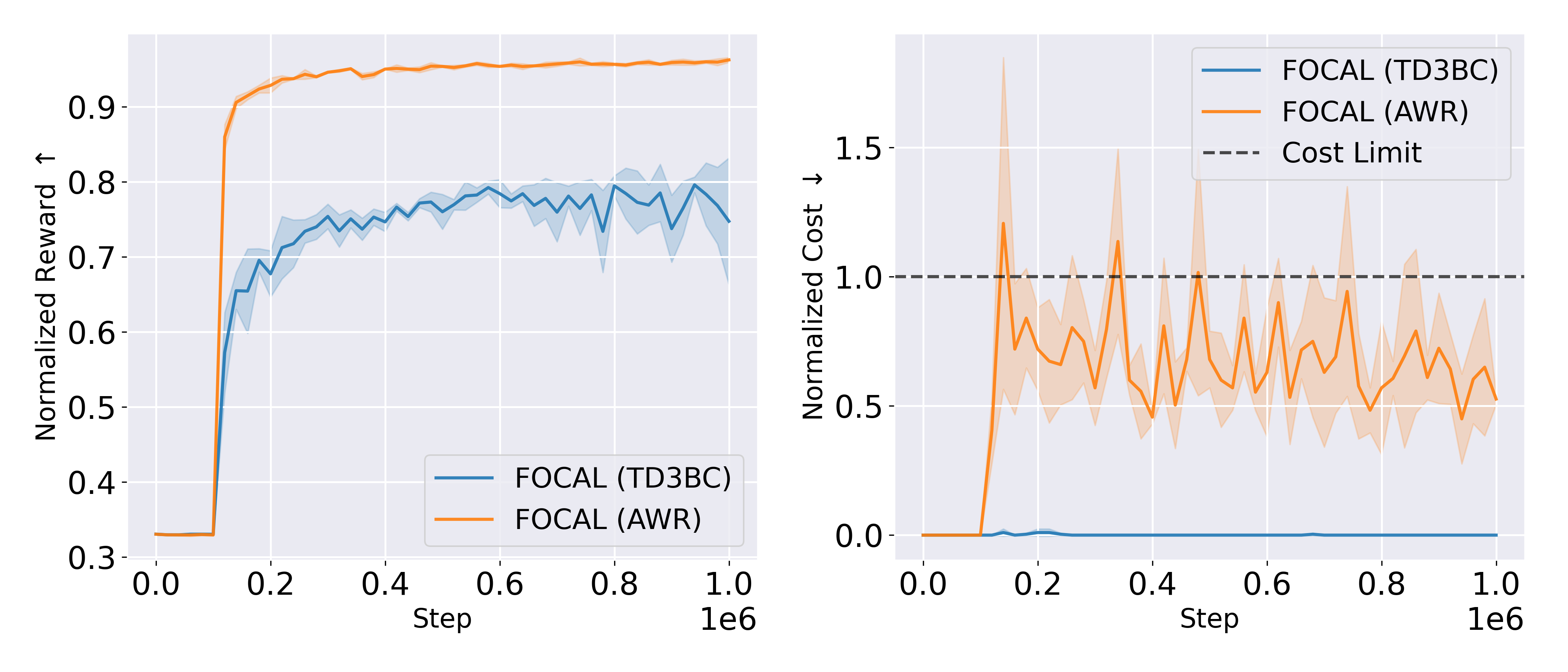}
        \caption{AntVelocity}
        \label{fig:ablation_awr_env1}
    \end{subfigure}
    \hfill
    \begin{subfigure}[b]{0.48\textwidth}
        \centering
        \includegraphics[width=\textwidth]{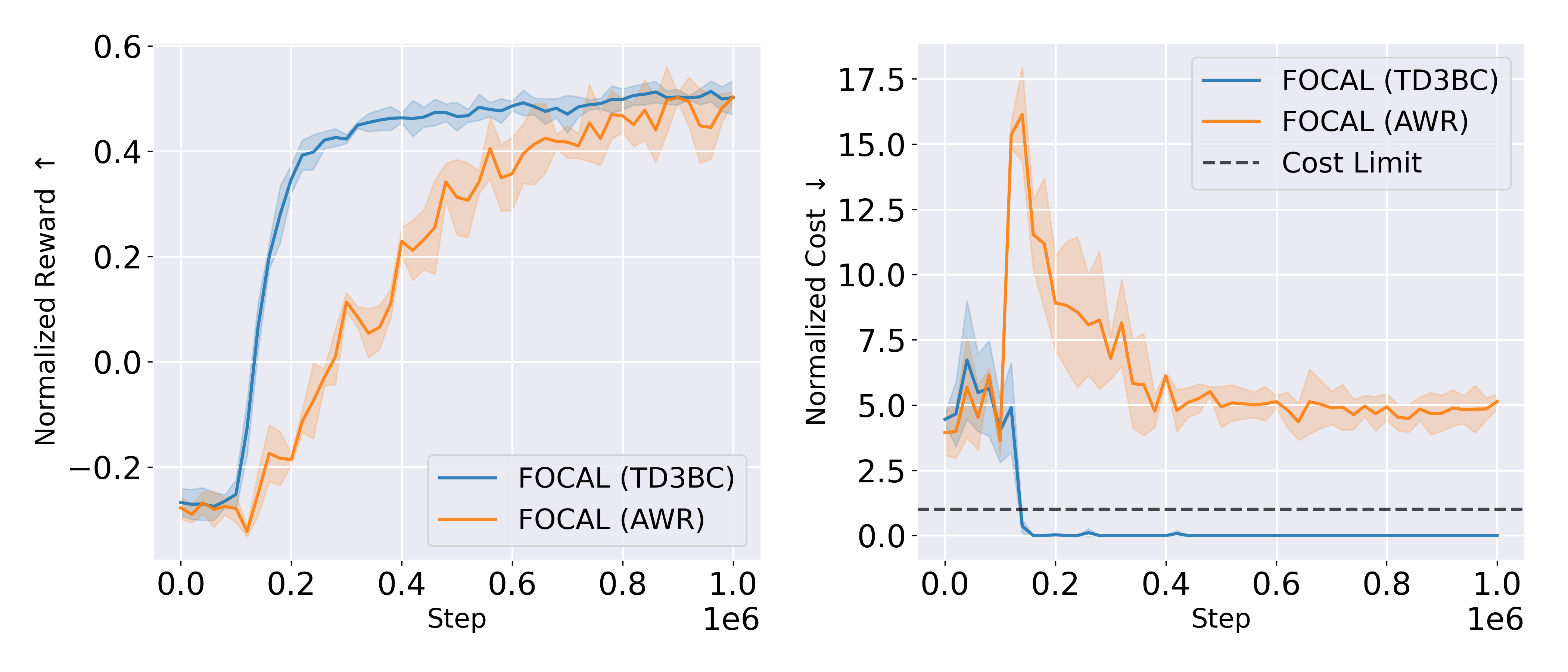}
        \caption{DroneCircle}
        \label{fig:ablation_awr_env2}
    \end{subfigure}
    \caption{\textbf{Ablation on the policy extraction objective.} Normalized return and
    cost during training for FOCAL with the first-order objective and with clipped AWR.
    Solid lines denote the mean over three training runs, and shaded regions
    denote one standard deviation.}
    \label{fig:ablation_awr}
\end{figure}

\textbf{First-order objective versus AWR.}
To evaluate the effect of the policy extraction objective, we replace the first-order
terms in Eq.~(\ref{eq:cvae_objective}) with AWR, keeping the feasibility critics, the masks,
and the inference procedure unchanged. The AWR variant uses the same exponential weight clipping as the
official FISOR implementation~\cite{zheng2024safe}. Figure~\ref{fig:ablation_awr} shows the training
dynamics on AntVelocity and DroneCircle. On both tasks, the AWR variant exhibits larger
variation in normalized cost across training and exceeds the cost threshold at several
points during training, whereas the first-order variant remains below the threshold.

\textbf{Adaptive latent scale.}
We compare the adaptive latent scale $\sigma_z(s)$ in Eq.~(\ref{eq:adaptive_variance}) with
fixed scales $\sigma_z \in \{1.0, 0.5, 0.1\}$, where the smallest fixed scale equals
$\sigma_{\min}$. All configurations use the same candidate filtering procedure
(Eq.~(\ref{eq:action_selection})) with $N = 16$ candidates, so that only the latent scale
differs. Table~\ref{tab:ablation_variance} reports the results. On AntVelocity and CarRun,
all configurations satisfy the cost threshold, and the adaptive scale obtains the highest
return. The adaptive scale satisfies the threshold on all four
tasks, with a return equal to or higher than that of every fixed scale that satisfies the
threshold. This suggests that adjusting the latent scale according to the estimated feasibility of
the state is more effective on these tasks than any single fixed scale.

\subsection{Inference Latency}
\label{sec:latency}

\begin{figure}[t]
    \centering
    % Left side: Table 2
    \begin{minipage}[c]{0.54\textwidth}
        \centering
        \captionof{table}{\textbf{Ablation on the latent scale at inference.} Normalized
        return and cost with fixed latent scales $\sigma_z$ and with the adaptive scale
        $\sigma_z(s)$. All configurations use candidate filtering with $N = 16$. Gray: unsafe
        ($C > 1$). Blue: safe configuration with the highest return.}
        \label{tab:ablation_variance}
        \resizebox{\linewidth}{!}{%
            \begin{tabular}{l c c c c c c c c}
                \toprule
                \multirow{2}{*}{\textbf{Task}} & \multicolumn{2}{c}{$\sigma_z = 1.0$} & \multicolumn{2}{c}{$\sigma_z = 0.5$} & \multicolumn{2}{c}{$\sigma_z = 0.1$} & \multicolumn{2}{c}{\textbf{Adaptive} $\sigma_z(s)$} \\
                \cmidrule(lr){2-3} \cmidrule(lr){4-5} \cmidrule(lr){6-7} \cmidrule(lr){8-9}
                & Return $\uparrow$ & Cost $\downarrow$ & Return $\uparrow$ & Cost $\downarrow$ & Return $\uparrow$ & Cost $\downarrow$ & Return $\uparrow$ & Cost $\downarrow$ \\
                \midrule
                AntVelocity & 0.76 & 0.00 & 0.77 & 0.00 & 0.74 & 0.00 & \textbf{\textcolor{blue}{0.81}} & \textbf{\textcolor{blue}{0.00}} \\
                CarGoal1 & \textcolor{gray}{0.62} & \textcolor{gray}{1.63} & \textcolor{gray}{0.62} & \textcolor{gray}{1.36} & \textcolor{gray}{0.59} & \textcolor{gray}{1.43} & \textbf{\textcolor{blue}{0.61}} & \textbf{\textcolor{blue}{0.80}} \\
                CarRun & 0.82 & 0.00 & 0.82 & 0.00 & 0.75 & 0.00 & \textbf{\textcolor{blue}{0.83}} & \textbf{\textcolor{blue}{0.00}} \\
                DroneRun & 0.56 & 0.80 & \textcolor{gray}{0.56} & \textcolor{gray}{1.26} & \textcolor{gray}{0.58} & \textcolor{gray}{1.33} & \textbf{\textcolor{blue}{0.58}} & \textbf{\textcolor{blue}{0.58}} \\
                \bottomrule
            \end{tabular}
        }
    \end{minipage}\hfill
    % Right side: Figure 2
    \begin{minipage}[c]{0.42\textwidth}
        \centering
        \includegraphics[width=\linewidth]{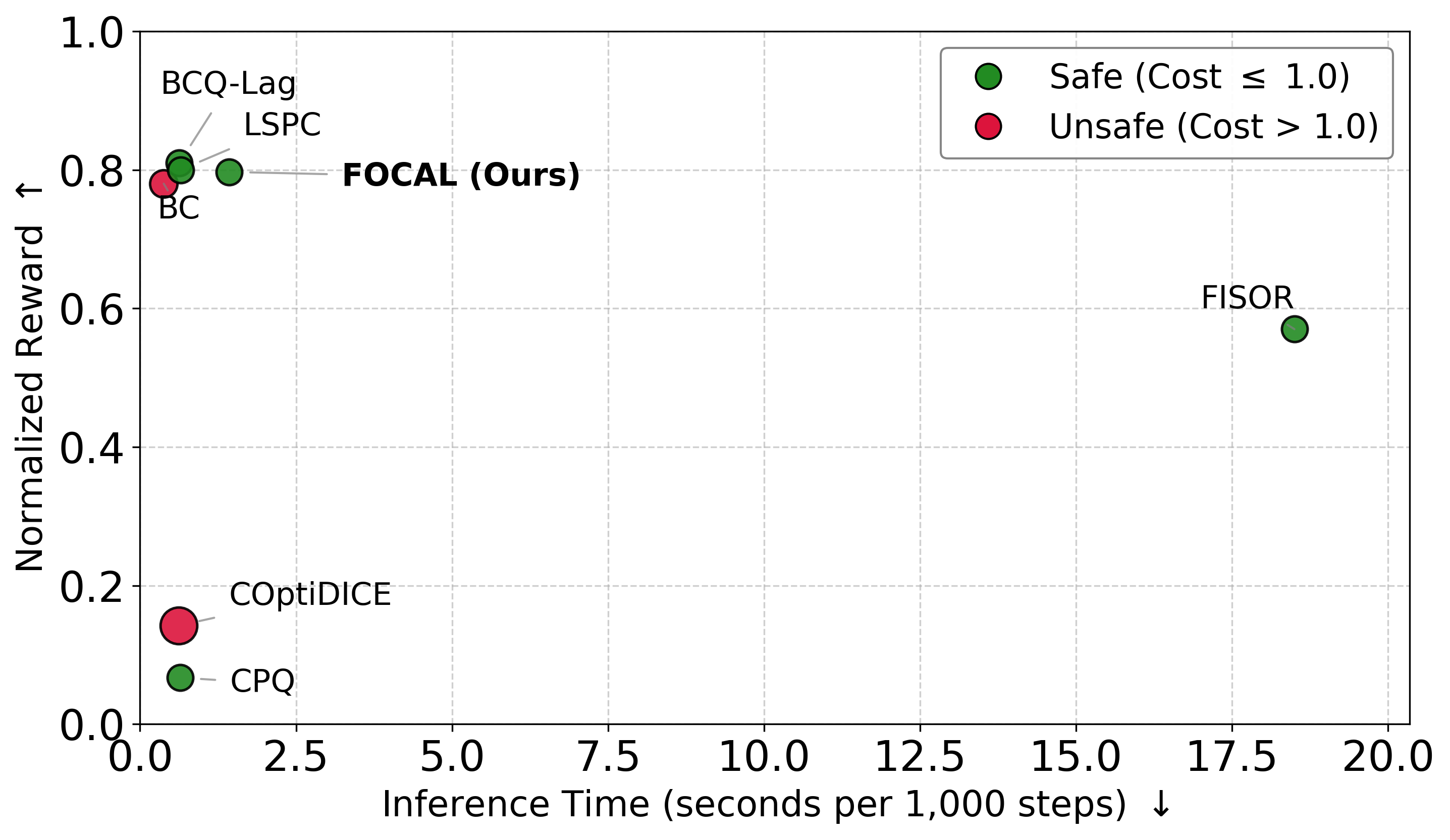}
        \caption{\textbf{Inference latency and performance.} Wall-clock time for $1{,}000$
        environment steps on Walker2d velocity with normalized return and cost.}
        \label{fig:inference_speed}
    \end{minipage}
\end{figure}

We compare the computational cost of policy execution by measuring the wall-clock time required to complete $1{,}000$ environment steps on the Walker2d velocity task, including both action generation and environment interaction. Figure~\ref{fig:inference_speed} reports this time together with the normalized return and cost of each method. Single-step baselines such as CPQ and COptiDICE require less than $1$ second, but exceed the cost threshold on this task. The diffusion-based FISOR satisfies the threshold but requires approximately $18$ seconds, while FOCAL satisfies the threshold and requires $1.43$ seconds. These results indicate that FOCAL retains the single-step inference of CVAE-based policies while satisfying the safety criterion on this task.

\section{Related Work}
\label{sec:related_work}

Offline RL learns policies from static datasets without environment interaction, avoiding the risks of online exploration \cite{levine2020offline}. Safe offline RL extends this setting to CMDPs, where policies must also satisfy safety constraints. A common approach is to constrain expected cumulative cost. BCQ-Lag combines behavior-constrained Q-learning with a Lagrangian cost constraint, CPQ \cite{xu2022constraints} penalizes unsafe or out-of-distribution actions, COptiDICE \cite{lee2022coptidice} optimizes the policy's stationary distribution under a cost constraint, and CDT \cite{liu2023constrained} conditions a sequence model on target returns and costs. However, expected-cost constraints bound aggregate violations without addressing state-wise safety.

Control-theoretic methods, such as Hamilton-Jacobi reachability, characterize states from which constraints can be satisfied at every future step \cite{bansal2017hamilton, fisac2019bridging}. RCRL \cite{yu2022reachability} uses reachability-based feasibility in online safe RL, while FISOR \cite{zheng2024safe} estimates feasible regions from offline data.

Offline datasets collected by multiple or suboptimal policies can exhibit multimodal action distributions, motivating generative policy models. Diffusion-based policies \cite{hansen2023idql, zheng2024safe} capture such distributions but require iterative denoising, increasing inference cost. Latent-variable models such as CVAEs generate actions in a single forward pass. In offline RL, CVAEs model behavior policies and constrain learned policies to their support through decoded-action perturbation \cite{fujimoto2019off} or latent-space optimization \cite{zhou2021plas}. LSPC \cite{koirala2024latent} extends CVAE-based modeling to latent safety constraints in safe offline RL.

AWR is commonly used for policy extraction when critic and policy learning are decoupled \cite{peng2019advantage, kostrikov2021offline, zheng2024safe, koirala2024latent}. It performs behavior cloning weighted by exponentiated advantage estimates. Although this preserves dataset support, exponential weighting amplifies critic errors and can make updates dominated by a few high-advantage transitions \cite{park2024value}. Behavior-regularized methods such as TD3+BC \cite{fujimoto2021minimalist} offer a first-order alternative that can better utilize learned value functions \cite{park2024value}. In safe offline RL, however, policy updates must account for both reward and hazard critics acting on the same action. Combining their gradients requires a trade-off weight, whereas FOCAL uses feasibility-based partitioning to select either the reward or hazard critic gradient for each transition. This enables first-order policy extraction with a single-step generative model, yielding update variance that depends linearly on critic estimation error (Theorem~\ref{theo:main_gradient_variance}).
\section{Conclusion}
\label{sec:conclusion}

We proposed FOCAL, a safe offline reinforcement learning method that trains a CVAE policy
with a feasibility-partitioned first-order objective based on learned reward and hazard
critics. For transitions in the estimated feasible region, the actor is updated in the
direction that increases the estimated return, and for transitions in the estimated
infeasible region, it is updated in the direction that decreases the estimated constraint
violation, while a reconstruction term keeps the generated actions close to the dataset
actions. At inference, FOCAL adjusts the scale of the latent prior according to the
estimated feasibility of the state and selects among candidate actions using the learned
critics. We analyzed how critic estimation error enters the policy update and showed that,
for unclipped AWR, the variance of the update depends exponentially on the advantage error, whereas for the proposed objective the variance
depends linearly on the critic error. On the DSRL benchmark, FOCAL satisfies
the constraints on all evaluated tasks and obtains competitive or higher returns compared to baselines.

The safety of FOCAL depends on the accuracy of the learned feasibility critics. These critics
are function approximators trained on a finite dataset, and our classification analysis
shows that a non-zero fraction of unsafe actions is classified as feasible at action
selection (Appendix~\ref{app:critic_calibration}). The accuracy of the critics further
depends on the coverage of the dataset near the constraint boundary, and limited coverage
can lead to overly conservative policies or to violations. FOCAL therefore does not provide
a state-wise safety guarantee at deployment. During training, the value of the hazard
critic at the training target is bounded by a quantity that increases with the reward step
size $\lambda_r$ (Proposition~\ref{prop:cvae_dpg_stationary}), so that a larger reward step
size permits larger deviations from feasible dataset actions. Future work could focus on uncertainty estimates for the hazard critic that could be used to penalize evaluations at
actions that are poorly covered by the data, and extending FOCAL to safe offline-to-online
fine-tuning could improve task performance through limited environment interaction.

\bibliographystyle{plainnat}
\bibliography{references}

\appendix
\newpage
\section{Theoretical Analysis of the FOCAL Objective}
\label{app:theoretical_proofs}

This section provides the derivations of the theoretical results on the FOCAL objective.
Section~\ref{subsubsec:stationary_point} derives the bound on the deviation of the training
target from the dataset action in Proposition~\ref{prop:cvae_dpg_stationary}.
Section~\ref{app:decoupling_proof} shows how the feasibility masks determine the critic
gradients received by the actor (Proposition~\ref{prop:decoupling}).
Section~\ref{subsubsec:variance_theorem} provides the proof of
Theorem~\ref{theo:main_gradient_variance}.

Throughout this section, the critics $\hat{Q}_r$, $\hat{Q}_h$, and $\hat{V}_h$ and the normalizing constants
$\Phi_r$ and $\Phi_h$ in Eq.~(\ref{eq:step_sizes}) are treated as constants with respect to
the actor parameters $\psi = (\theta, \phi)$, as in the implementation. Consequently, the
step sizes $\lambda_r$ and $\lambda_h$ do not depend on $\psi$, and the masks
$\mathcal{M}_{\mathrm{feas}}(s,a)$, $\mathcal{M}_{\mathrm{infeas}}(s,a)$, and
$\mathcal{M}_{\mathrm{valid}}(s,a)$ in Eq.~(\ref{eq:masks}) depend only on the dataset
transition $(s,a)$. We write $\mathcal{D}_{\mathrm{valid}} := \{(s,a) \in \mathcal{D} :
\mathcal{M}_{\mathrm{valid}}(s,a) = 1\}$ for the set of transitions that contribute to the
objective.

% ---------------------------------------------------------------------------
\subsection{Bounded Action Deviation}
\label{subsubsec:stationary_point}

For a transition $(s,a) \in \mathcal{D}_{\mathrm{valid}}$, the value-dependent part of the
objective in Eq.~(\ref{eq:cvae_objective}) is, as a function of the reconstructed action
$\hat a$,
\begin{align}
\ell_{\mathrm{feas}}(\hat a) &= \lVert a - \hat a \rVert_2^2 - \lambda_r\, \hat{Q}_r(s,\hat a)
&& \text{if } \mathcal{M}_{\mathrm{feas}}(s,a) = 1,
\label{eq:region_losses}
\\
\ell_{\mathrm{infeas}}(\hat a) &= \lVert a - \hat a \rVert_2^2 + \lambda_h\, \hat{Q}_h(s,\hat a)
&& \text{if } \mathcal{M}_{\mathrm{infeas}}(s,a) = 1.
\nonumber
\end{align}
We refer to a stationary point $\hat a^*$ of the applicable loss as the training target for
$(s,a)$. Below, $\nabla_{\hat a} \hat{Q}(s,\hat a^*)$ denotes the gradient of a critic $\hat{Q}(s,\cdot)$ with
respect to its action argument, evaluated at $\hat a^*$.

\medskip
\noindent\textbf{Proposition~\ref{prop:cvae_dpg_stationary} (restated).}
\textit{Let Assumption~\ref{assum:critic_lipschitz} hold, and let $(s,a) \in \mathcal{D}_{\mathrm{valid}}$.
Let $\hat a^*$ be a stationary point, in the interior of the action space, of the
region-specific loss in Eq.~(\ref{eq:region_losses}) that applies to $(s,a)$. Then}
\begin{align}
\text{(Feasible)} \quad
&\hat a^* = a + \frac{\lambda_r}{2}\, \nabla_{\hat a} \hat{Q}_r(s,\hat a^*),
\qquad
\lVert \hat a^* - a \rVert_2 \le \frac{\lambda_r L_{Q_r}}{2},
\label{eq:app_prop_feas}
\\
&\hat{Q}_h(s,\hat a^*) \le \hat{Q}_h(s,a) + \frac{\lambda_r L_{Q_r} L_{Q_h}}{2}
\le \frac{\lambda_r L_{Q_r} L_{Q_h}}{2},
\label{eq:app_prop_hazard}
\\
\text{(Infeasible)} \quad
&\hat a^* = a - \frac{\lambda_h}{2}\, \nabla_{\hat a} \hat{Q}_h(s,\hat a^*),
\qquad
\lVert \hat a^* - a \rVert_2 \le \frac{\lambda_h L_{Q_h}}{2}.
\label{eq:app_prop_infeas}
\end{align}
\textit{The bounds characterize the training target defined by the region-specific loss. They do not apply to actions decoded at test time.}

\begin{proof}
\emph{Feasible case.} Since $\hat a^*$ is a
stationary point of $\ell_{\mathrm{feas}}$ in the interior of the action space and
$\lambda_r$ does not depend on $\hat a$, the first-order condition is
\[
\nabla_{\hat a}\, \ell_{\mathrm{feas}}(\hat a^*)
= -2\,(a - \hat a^*) - \lambda_r\, \nabla_{\hat a} \hat{Q}_r(s,\hat a^*) = 0.
\]
Rearranging gives the fixed-point equation
$\hat a^* = a + \tfrac{\lambda_r}{2}\, \nabla_{\hat a} \hat{Q}_r(s,\hat a^*)$. Taking the
$\ell_2$ norm of both sides,
$\lVert \hat a^* - a \rVert_2 = \tfrac{\lambda_r}{2}\, \lVert \nabla_{\hat a} \hat{Q}_r(s,\hat a^*) \rVert_2$.
Since $\hat{Q}_r(s,\cdot)$ is $L_{Q_r}$-Lipschitz, its gradient satisfies
$\lVert \nabla_{\hat a} \hat{Q}_r(s,\hat a^*) \rVert_2 \le L_{Q_r}$ which gives
the deviation bound in Eq.~(\ref{eq:app_prop_feas}).

For the hazard value, since $\hat{Q}_h(s,\cdot)$ is $L_{Q_h}$-Lipschitz,
\[
\hat{Q}_h(s,\hat a^*) \le \hat{Q}_h(s,a) + L_{Q_h}\, \lVert \hat a^* - a \rVert_2
\le \hat{Q}_h(s,a) + \frac{\lambda_r L_{Q_r} L_{Q_h}}{2}.
\]
Since $\mathcal{M}_{\mathrm{feas}}(s,a) = 1$ implies $\hat{Q}_h(s,a) \le 0$, the second inequality
in Eq.~(\ref{eq:app_prop_hazard}) follows.

\emph{Infeasible case.} The first-order condition
for $\ell_{\mathrm{infeas}}$ is
\[
\nabla_{\hat a}\, \ell_{\mathrm{infeas}}(\hat a^*)
= -2\,(a - \hat a^*) + \lambda_h\, \nabla_{\hat a} \hat{Q}_h(s,\hat a^*) = 0,
\]
which gives $\hat a^* = a - \tfrac{\lambda_h}{2}\, \nabla_{\hat a} \hat{Q}_h(s,\hat a^*)$. Taking
the $\ell_2$ norm and using
$\lVert \nabla_{\hat a} \hat{Q}_h(s,\hat a^*) \rVert_2 \le L_{Q_h}$ gives the bound in
Eq.~(\ref{eq:app_prop_infeas}).
\end{proof}
% ---------------------------------------------------------------------------

\subsection{Feasibility-Partitioned Gradients}
\label{app:decoupling_proof}

We now characterize the gradient of the objective in Eq.~(\ref{eq:cvae_objective}) with
respect to the actor parameters. We use the reparameterization
$z = m_\phi(s,a) + s_\phi(s,a) \odot \varepsilon$ with $\varepsilon \sim \mathcal{N}(0, I)$,
where $m_\phi$ and $s_\phi$ denote the mean and standard deviation of the encoder, so that
the reconstructed action $\hat a = D_\theta(s,z)$ is a differentiable function of $\psi$ for
fixed $(s,a,\varepsilon)$. For a fixed $(s,a,\varepsilon)$, the per-sample loss is
\begin{equation}
\label{eq:per_sample_loss}
\begin{split}
\ell(\psi; s,a,\varepsilon)
={}& \mathcal{M}_{\mathrm{feas}}(s,a)
\big( \lVert a - \hat a \rVert_2^2 - \lambda_r\, \hat{Q}_r(s,\hat a) \big)
+ \mathcal{M}_{\mathrm{infeas}}(s,a)
\big( \lVert a - \hat a \rVert_2^2 + \lambda_h\, \hat{Q}_h(s,\hat a) \big)
\\
&+ \mathcal{M}_{\mathrm{valid}}(s,a)\, \beta\,
D_{\mathrm{KL}}\big( q_\phi(z \mid s,a) \,\Vert\, \mathcal{N}(0,I) \big),
\end{split}
\end{equation}
so that $\mathcal{L}_{\mathrm{CVAE}}(\theta,\phi) = \mathbb{E}_{(s,a)\sim\mathcal{D},\,
\varepsilon}\big[ \ell(\psi; s,a,\varepsilon) \big]$. Let
$J_\psi := \partial \hat a / \partial \psi$ denote the Jacobian of the reconstructed action
with respect to the actor parameters.

\begin{proposition}[Feasibility-partitioned gradients]
\label{prop:decoupling}
For a transition $(s,a)$ and noise $\varepsilon$, the gradient of the
value-dependent term of Eq.~(\ref{eq:per_sample_loss}) with respect to $\psi$ is
\begin{equation}
\label{eq:per_sample_gradient}
\begin{cases}
-\lambda_r\, J_\psi^{\top}\, \nabla_{\hat a} \hat{Q}_r(s,\hat a), & \text{if } \mathcal{M}_{\mathrm{feas}}(s,a) = 1,\\[2pt]
\phantom{-}\lambda_h\, J_\psi^{\top}\, \nabla_{\hat a} \hat{Q}_h(s,\hat a), & \text{if } \mathcal{M}_{\mathrm{infeas}}(s,a) = 1,\\[2pt]
\phantom{-}0, & \text{if } \mathcal{M}_{\mathrm{valid}}(s,a) = 0.
\end{cases}
\end{equation}
Consequently, each valid transition contributes the action gradient of exactly one of the
critics $\hat{Q}_r$ and $\hat{Q}_h$.
\end{proposition}

Proposition~\ref{prop:decoupling} states that no single transition combines the gradients of
the reward and hazard critics. Since the parameters $\psi$ are shared, the mini-batch
gradient still aggregates contributions from both regions.

% ---------------------------------------------------------------------------
\subsection{Dependence of the Policy Update on Critic Estimation Error}
\label{subsubsec:variance_theorem}

This section provides the derivation of Theorem~\ref{theo:main_gradient_variance} under
Assumptions~\ref{assum:critic_error}--\ref{assum:decoder}. For a dataset pair $(s,a)$, we
abbreviate $\tilde\pi := \tilde\pi_\psi(\cdot \mid s,a)$ and
$u(\hat a) := \nabla_\psi \log \tilde\pi_\psi(\hat a \mid s,a)$, so that
$\operatorname{tr} F(\psi;s,a) = \mathbb{E}_{\hat a \sim \tilde\pi}\big[ \lVert u(\hat a) \rVert_2^2 \big]$.

\medskip
\noindent\textbf{Theorem~\ref{theo:main_gradient_variance} (restated).}
\textit{Fix a dataset transition $(s,a)$, and let $\mathbb{E}_\epsilon$ denote expectation over critic training randomness.}

\textit{\textbf{(i) AWR.} We analyze AWR with the unclipped exponential weight. Let $\tau > 0$ be the temperature parameter, and let $v := \nabla_\psi \log \pi_\psi(a \mid s)$. Define $g_{\mathrm{AWR}} := \exp\big(\hat A(s,a)/\tau\big)\, v$ and $g^*_{\mathrm{AWR}} := \exp\big(A^*(s,a)/\tau\big)\, v$. Under Assumption~\ref{assum:adv_error}, with $A^* = A^*(s,a)$, $\mu_A = \mu_A(s,a)$, and $\sigma_A = \sigma_A(s,a)$:
\begin{align}
\big\lVert \mathbb{E}_\epsilon[g_{\mathrm{AWR}}] - g^*_{\mathrm{AWR}} \big\rVert_2
&= \lVert v \rVert_2\, e^{A^*/\tau}\, \Big\lvert e^{\mu_A/\tau + \sigma_A^2/(2\tau^2)} - 1 \Big\rvert, \label{eq:app_awr_bias} \\
\mathbb{E}_\epsilon\Big[ \big\lVert g_{\mathrm{AWR}} - \mathbb{E}_\epsilon[g_{\mathrm{AWR}}] \big\rVert_2^2 \Big]
&= \lVert v \rVert_2^2\, e^{2A^*/\tau}\, e^{2\mu_A/\tau + \sigma_A^2/\tau^2} \Big( e^{\sigma_A^2/\tau^2} - 1 \Big). \label{eq:app_awr_var}
\end{align}}

\textit{\textbf{(ii) FOCAL.} Let $g_{\mathrm{FOCAL}} := \lambda\, \nabla_\psi\, \mathbb{E}_{\hat a \sim \tilde\pi_\psi(\cdot \mid s,a)}\big[ \hat Q(s,\hat a) \big]$ with error-free counterpart $g^*_{\mathrm{FOCAL}} := \lambda\, \nabla_\psi\, \mathbb{E}_{\hat a \sim \tilde\pi_\psi(\cdot \mid s,a)}\big[ Q^*(s,\hat a) \big]$. Under Assumptions~\ref{assum:critic_error} and~\ref{assum:decoder}:
\begin{align}
\big\lVert \mathbb{E}_\epsilon[g_{\mathrm{FOCAL}}] - g^*_{\mathrm{FOCAL}} \big\rVert_2
&\le \lambda\, \sqrt{\operatorname{tr} F(\psi;s,a)}\, \sqrt{\mathrm{Var}_{\hat a \sim \tilde\pi_\psi}\big[ \mu(s,\hat a) \big]}, \label{eq:app_focal_bias} \\
\mathbb{E}_\epsilon\Big[ \big\lVert g_{\mathrm{FOCAL}} - \mathbb{E}_\epsilon[g_{\mathrm{FOCAL}}] \big\rVert_2^2 \Big]
&\le \lambda^2\, \operatorname{tr} F(\psi;s,a)\, \mathbb{E}_\epsilon\Big[ \mathrm{Var}_{\hat a \sim \tilde\pi_\psi}\big[ \xi(s,\hat a) \big] \Big] \nonumber \\
&\le \lambda^2\, \operatorname{tr} F(\psi;s,a)\, \mathbb{E}_{\hat a \sim \tilde\pi_\psi}\big[ \sigma^2(s,\hat a) \big], \label{eq:app_focal_var}
\end{align}
where the trace of the Fisher information is bounded by a decoder term and the Fisher
information of the encoder:
\begin{equation}
\operatorname{tr} F(\psi;s,a) \le \frac{d\, L_\theta^2}{\sigma_{\mathrm{dec}}^2}
+ \operatorname{tr} F_q(\phi;s,a). \label{eq:app_fisher}
\end{equation}
The same bounds hold for the infeasible branch with $(\hat Q, \lambda)$ replaced by $(\hat Q_h, \lambda_h)$.}

\begin{proof}
\textbf{Part (i): AWR.} Under Assumption~\ref{assum:adv_error}, write $\hat A = A^* + \epsilon_A$ and $X := e^{\epsilon_A/\tau}$, so that $g_{\mathrm{AWR}} = X\, g^*_{\mathrm{AWR}}$. Since $\epsilon_A \sim \mathcal{N}(\mu_A, \sigma_A^2)$, evaluating the Gaussian moment-generating function $\mathbb{E}[e^{c\,\epsilon_A}] = \exp(c\mu_A + c^2\sigma_A^2/2)$ at $c \in \{1/\tau, 2/\tau\}$ yields:
\[
\mathbb{E}_\epsilon[X] = e^{\mu_A/\tau + \sigma_A^2/(2\tau^2)}, \quad 
\mathbb{E}_\epsilon[X^2] = e^{2\mu_A/\tau + 2\sigma_A^2/\tau^2}, \quad
\mathrm{Var}_\epsilon[X] = e^{2\mu_A/\tau + \sigma_A^2/\tau^2}\big(e^{\sigma_A^2/\tau^2} - 1\big).
\]
Eqs.~(\ref{eq:app_awr_bias}) and~(\ref{eq:app_awr_var}) follow directly by computing $\lVert \mathbb{E}_\epsilon[g_{\mathrm{AWR}}] - g^*_{\mathrm{AWR}} \rVert_2 = \lvert \mathbb{E}_\epsilon[X] - 1 \rvert \lVert g^*_{\mathrm{AWR}} \rVert_2$ and $\mathbb{E}_\epsilon\lVert g_{\mathrm{AWR}} - \mathbb{E}_\epsilon[g_{\mathrm{AWR}}] \rVert_2^2 = \lVert g^*_{\mathrm{AWR}} \rVert_2^2\, \mathrm{Var}_\epsilon[X]$.

\textbf{Part (ii): FOCAL.} Abbreviate $\tilde\pi := \tilde\pi_\psi(\cdot \mid s,a)$.

\emph{Step 1 (Score-function representation).} By Assumption~\ref{assum:decoder}, differentiation with respect to $\psi$ interchanges with integration over $\tilde\pi$. Applying the log-derivative identity $\nabla_\psi \tilde\pi = \tilde\pi \nabla_\psi \log \tilde\pi = \tilde\pi u$, the policy gradient for any integrable function $f$ satisfies:
\begin{equation}
\label{eq:app_score_identity}
\nabla_\psi \mathbb{E}_{\hat a \sim \tilde\pi}[f(\hat a)] = \int f(\hat a)\, \nabla_\psi \tilde\pi(\hat a \mid s,a)\, d\hat a = \mathbb{E}_{\hat a \sim \tilde\pi}\big[ f(\hat a)\, u(\hat a) \big].
\end{equation}
Subtracting $g^*_{\mathrm{FOCAL}}$ from $g_{\mathrm{FOCAL}}$ and defining $\Delta := g_{\mathrm{FOCAL}} - g^*_{\mathrm{FOCAL}}$ yields:
\begin{equation}
\label{eq:app_delta}
\Delta = \lambda\, \mathbb{E}_{\hat a \sim \tilde\pi}\big[ \epsilon_Q(s,\hat a)\, u(\hat a) \big].
\end{equation}
 
\emph{Step 2 (Baseline invariance).} Since $\int \tilde\pi(\hat a \mid s,a)\, d\hat a = 1$ for all $\psi$, differentiating both sides gives $\mathbb{E}_{\hat a \sim \tilde\pi}[u(\hat a)] = \nabla_\psi (1) = 0$. Hence, for any baseline $b$ that is constant with respect to $\hat a$:
\begin{equation}
\label{eq:app_baseline}
\mathbb{E}_{\hat a \sim \tilde\pi}\big[ \epsilon_Q(s,\hat a)\, u(\hat a) \big] = \mathbb{E}_{\hat a \sim \tilde\pi}\big[ \big(\epsilon_Q(s,\hat a) - b\big)\, u(\hat a) \big].
\end{equation}
 
\emph{Step 3 (Cauchy--Schwarz separation).} For any scalar function $Y(\hat a)$ and vector $u(\hat a)$, Jensen's inequality and the scalar Cauchy--Schwarz inequality imply:
\begin{equation}
\label{eq:app_cs}
\big\lVert \mathbb{E}_{\hat a \sim \tilde\pi}\big[ Y(\hat a)\, u(\hat a) \big] \big\rVert_2 \le \mathbb{E}_{\hat a \sim \tilde\pi}\big[ |Y(\hat a)|\, \lVert u(\hat a) \rVert_2 \big] \le \sqrt{\mathbb{E}_{\hat a \sim \tilde\pi}[Y(\hat a)^2]}\; \sqrt{\operatorname{tr} F(\psi;s,a)},
\end{equation}
using $\mathbb{E}_{\hat a \sim \tilde\pi}[\lVert u(\hat a) \rVert_2^2] = \operatorname{tr} F(\psi;s,a)$.
 
\emph{Step 4 (Bias bound).} Taking the expectation over critic error in Eq.~(\ref{eq:app_delta}) and swapping expectations (Fubini) yields $\mathbb{E}_\epsilon[\Delta] = \lambda\, \mathbb{E}_{\hat a \sim \tilde\pi}[\mu(s,\hat a)\, u(\hat a)]$. Applying Eq.~(\ref{eq:app_baseline}) with $b = \bar\mu := \mathbb{E}_{\hat a \sim \tilde\pi}[\mu(s,\hat a)]$ and Eq.~(\ref{eq:app_cs}) with $Y = \mu(s,\cdot) - \bar\mu$ gives Eq.~(\ref{eq:app_focal_bias}):
\[
\lVert \mathbb{E}_\epsilon[\Delta] \rVert_2 \le \lambda\, \sqrt{\operatorname{tr} F(\psi;s,a)}\; \sqrt{\mathrm{Var}_{\hat a \sim \tilde\pi}[\mu(s,\hat a)]}.
\]
 
\emph{Step 5 (Variance bound).} Subtracting $\mathbb{E}_\epsilon[\Delta]$ from Eq.~(\ref{eq:app_delta}) gives $\Delta - \mathbb{E}_\epsilon[\Delta] = \lambda\, \mathbb{E}_{\hat a \sim \tilde\pi}[\xi(s,\hat a)\, u(\hat a)]$. For any fixed critic realization, $\bar\xi := \mathbb{E}_{\hat a \sim \tilde\pi}[\xi(s,\hat a)]$ is constant with respect to $\hat a$. Setting $b = \bar\xi$ in Eq.~(\ref{eq:app_baseline}) and applying Eq.~(\ref{eq:app_cs}) with $Y = \xi(s,\cdot) - \bar\xi$:
\[
\lVert \Delta - \mathbb{E}_\epsilon[\Delta] \rVert_2^2 \le \lambda^2\, \operatorname{tr} F(\psi;s,a)\, \mathrm{Var}_{\hat a \sim \tilde\pi}\big[\xi(s,\hat a)\big].
\]
Taking expectation $\mathbb{E}_\epsilon$, applying Tonelli's theorem (since $\xi^2 \ge 0$), and using $\mathbb{E}_\epsilon[\xi^2] = \sigma^2$:
\[
\mathbb{E}_\epsilon\big[ \mathrm{Var}_{\hat a \sim \tilde\pi}[\xi] \big] = \mathbb{E}_{\hat a \sim \tilde\pi}\big[ \sigma^2(s,\hat a) \big] - \mathbb{E}_\epsilon[\bar\xi^{\,2}] \le \mathbb{E}_{\hat a \sim \tilde\pi}\big[ \sigma^2(s,\hat a) \big],
\]
since $\mathbb{E}_\epsilon[\bar\xi^{\,2}] \ge 0$. This establishes Eq.~(\ref{eq:app_focal_var}).
 
\emph{Step 6 (Fisher information bound).} The bounds in Steps 4 and 5 are proportional to
$\operatorname{tr} F(\psi;s,a)$, which measures how strongly the distribution of generated
actions changes with the actor parameters. In this step, we bound it in
terms of the action dimension $d$, the decoder Lipschitz constant $L_\theta$, the decoder
standard deviation $\sigma_{\mathrm{dec}}$, and the Fisher information of the encoder.

The score decomposes into the blocks $u = (\nabla_\theta \log\tilde\pi, \nabla_\phi \log\tilde\pi)$, 
so that $\operatorname{tr} F = \mathbb{E}\lVert \nabla_\theta \log\tilde\pi \rVert_2^2 + \mathbb{E}\lVert \nabla_\phi \log\tilde\pi \rVert_2^2$. 
We bound the two blocks separately. For the mixture density $\tilde\pi_\psi(\hat a \mid s,a) = \int p_\theta(\hat a \mid s,z)\, q_\phi(z \mid s,a)\, dz$, let $r(z \mid \hat a, s, a) \propto p_\theta(\hat a \mid s,z)\, q_\phi(z \mid s,a)$ denote the posterior over $z$. Since $\theta$ enters only through $p_\theta$,
   $\nabla_\theta \log \tilde\pi_\psi(\hat a \mid s,a)
   = \mathbb{E}_{z \sim r(\cdot \mid \hat a, s, a)}\big[ \nabla_\theta \log p_\theta(\hat a \mid s,z) \big]$,
   and by Jensen's inequality:
\[
\mathbb{E}_{\hat a \sim \tilde\pi}\big\lVert \nabla_\theta \log \tilde\pi_\psi(\hat a \mid s,a) \big\rVert_2^2 \le \mathbb{E}_{z \sim q_\phi, \hat a \sim p_\theta}\big\lVert \nabla_\theta \log p_\theta(\hat a \mid s,z) \big\rVert_2^2.
\]
For $p_\theta(\hat a \mid s,z) = \mathcal{N}\big(D_\theta(s,z), \sigma_{\mathrm{dec}}^2 I_d\big)$ with $\hat a = D_\theta(s,z) + \sigma_{\mathrm{dec}}\eta$, we have $\nabla_\theta \log p_\theta = J_\theta^{\top}\eta/\sigma_{\mathrm{dec}}$. Since $J_\theta$ has $d$ rows, $\lVert J_\theta \rVert_F^2 \le d\,\lVert J_\theta \rVert_2^2 \le d\,L_\theta^2$. Hence:
\[
\mathbb{E}_{\eta}\big\lVert J_\theta^{\top}\eta \big\rVert_2^2 / \sigma_{\mathrm{dec}}^2 = \lVert J_\theta \rVert_F^2 / \sigma_{\mathrm{dec}}^2 \le d\, L_\theta^2 / \sigma_{\mathrm{dec}}^2.
\]
Applying an identical Jensen argument to the encoder block yields $\mathbb{E}\lVert \nabla_\phi \log \tilde\pi \rVert_2^2 \le \operatorname{tr} F_q(\phi;s,a)$, establishing Eq.~(\ref{eq:app_fisher}). The encoder term $\operatorname{tr} F_q(\phi;s,a)$ depends only on the encoder and is
independent of the critic error, we do not bound it further in general.
 
\emph{Infeasible branch.} The derivation of part~(ii) uses only
Assumption~\ref{assum:critic_error} for the critic and Assumption~\ref{assum:decoder} for the
actor. It therefore applies without change to the hazard critic $\hat Q_h$ with step size
$\lambda_h$.

\end{proof}

\begin{figure}[t!]
    \centering
    % First Subfigure: Learning Curves
    \begin{subfigure}[b]{0.48\textwidth}
        \centering
        % Replace with your actual learning curve PDF path
        \includegraphics[width=\textwidth]{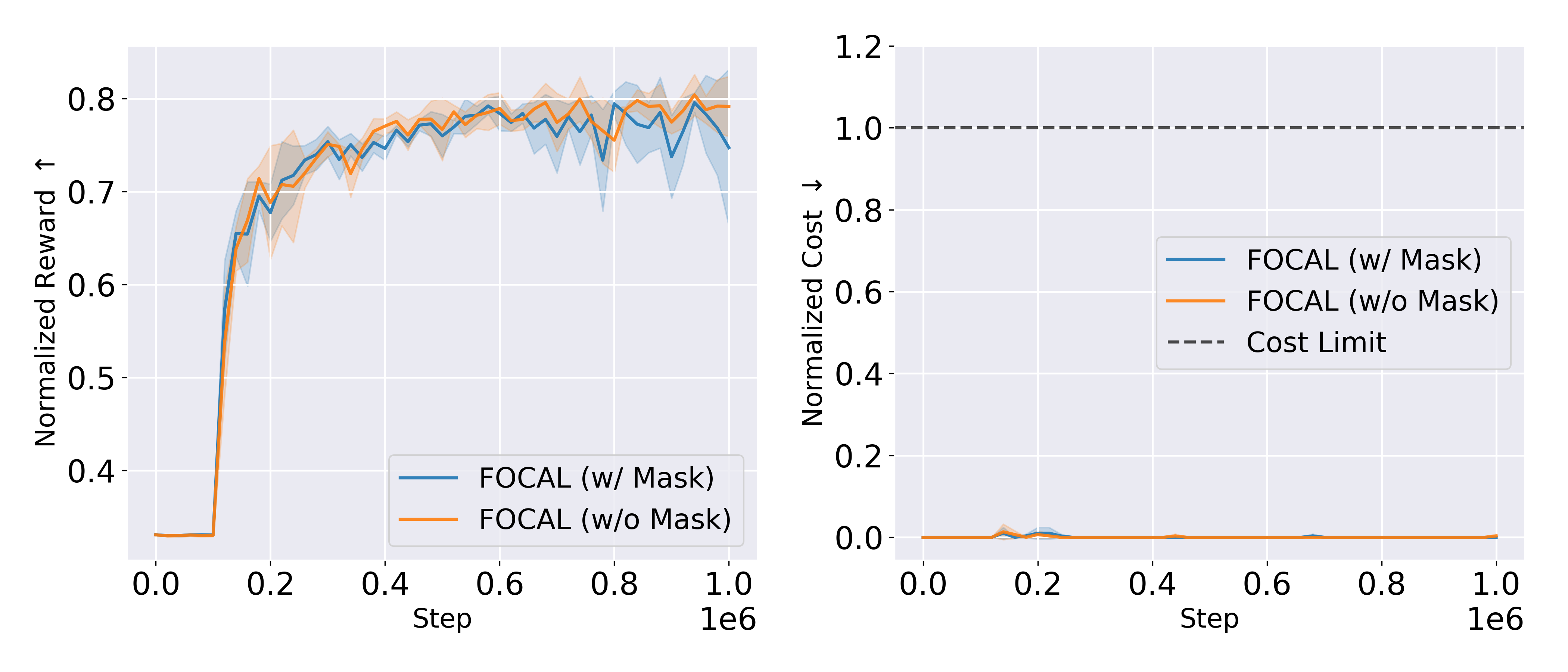}
        \caption{Learning Dynamics}
        \label{fig:ablation_case3_curves}
    \end{subfigure}
    \hfill
    % Second Subfigure: Contradiction Ratio
    \begin{subfigure}[b]{0.48\textwidth}
        \centering
        % Replace with your actual W&B plot PDF path
        \includegraphics[width=\textwidth]{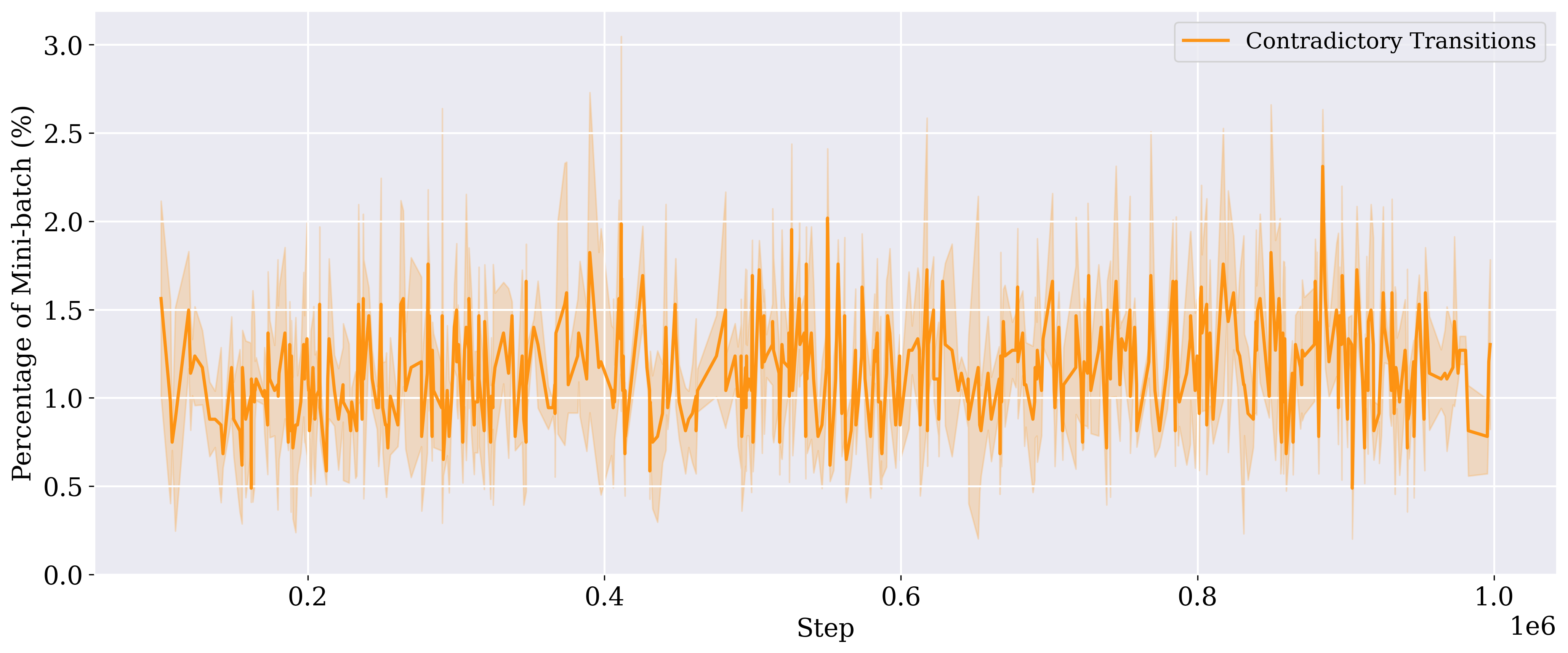}
        \caption{Contradiction Ratio}
        \label{fig:ablation_case3_ratio}
    \end{subfigure}
    
    \vspace{2mm}
    \caption{\textbf{Ablation on the Feasibility Contradiction Mask.} \textit{(a)} Training curves comparing FOCAL with and without the contradiction mask. \textit{(b)} The tracked proportion of training transitions satisfying the contradiction condition ($\hat V_h > 0 \land \hat Q_h \le 0$).}
    \label{fig:ablation_case3}
\end{figure}

\section{Additional Ablations}
\label{app:additional_ablations}

\subsection{Effect of the Contradiction Mask}
\label{app:contradiction_mask}

The masks in Eq.~(\ref{eq:masks}) exclude transitions that satisfy
$\hat V_h(s) > 0 \wedge \hat Q_h(s,a) \le 0$. Since $\hat V_h$ is trained to approximate a lower expectile
of $\hat Q_h(s,\cdot)$ over the dataset actions, a state estimated to be infeasible should not
admit a dataset action with a non-positive hazard value. Such transitions therefore
indicate an inconsistency between the two learned critics, which can arise from function
approximation error. FOCAL excludes them from the objective.

To assess the effect of this exclusion, we train FOCAL with and without the contradiction
mask on AntVelocity. As shown in Figure~\ref{fig:ablation_case3}(a), the two variants
obtain similar normalized return and cost. Figure~\ref{fig:ablation_case3}(b) reports the
fraction of transitions in each mini-batch that satisfy the contradiction condition and this
fraction remains below $5\%$ throughout training. The exclusion therefore affects only a
small part of each mini-batch on this task, which is consistent with the similar
performance of the two variants. We retain the mask because it prevents policy updates
from relying on inconsistent critic estimates, which may occur more frequently when
function approximation error is larger or data coverage is more limited.

% ---------------------------------------------------------------------------
\subsection{Effect of the Number of Candidate Actions}
\label{app:action_sampling_ablation}

At inference, FOCAL decodes $N$ candidate actions, retains those estimated to be feasible
by $\hat Q_h$, and executes the retained candidate with the highest estimated reward
(Eq.~(\ref{eq:action_selection})). We evaluate $N \in \{1, 4, 16, 64\}$ on two
Safety-Gymnasium tasks (Walker2d and Hopper velocity) and three Bullet-Safety-Gym tasks
(AntRun, BallCircle, and BallRun). Table~\ref{tab:ablation_sampling} reports the results.

All configurations satisfy the cost threshold on all five tasks. On the Safety-Gymnasium
tasks, the normalized cost generally decreases as $N$ increases, since a larger candidate
set is more likely to contain actions that $\hat Q_h$ classifies as feasible. The effect on
return depends on the task. On Hopper, increasing $N$ from $1$ to $16$ increases the
normalized return from $0.50$ to $0.83$ and decreases the normalized cost from $0.40$ to
$0.00$. On Walker2d, increasing $N$ decreases the normalized cost from $0.50$ to $0.31$,
while the normalized return decreases slightly from $0.75$ to $0.69$, which suggests that
a larger candidate set can favor actions with lower estimated hazard at the expense of
return. On the three Bullet-Safety-Gym tasks, the normalized cost is zero for all values of
$N$, and $N = 16$ obtains the highest return. Based on these results, we use $N = 16$ in
all experiments.

% ---------------------------------------------------------------------------
\subsection{Empirical Distribution of Feasibility Values}
\label{sec:vh_scaling_ablation}

The adaptive latent scale in Eq.~(\ref{eq:adaptive_variance}),
$\sigma_z(s) = \mathrm{clip}(-\hat V_h(s), \sigma_{\min}, 1)$, varies with the state only when
$\hat V_h(s) \in (-1, -\sigma_{\min})$ and outside this interval, it equals one of the clipping
limits. To examine how often the
scale varies in practice, we record $\hat V_h(s)$ at all states visited during $20$ evaluation
episodes on Hopper velocity and DroneRun with $\sigma_{\min} = 0.1$.

\begin{figure}[t]
    \centering
    \begin{subfigure}{0.48\textwidth}
        \centering
        \includegraphics[width=\linewidth]{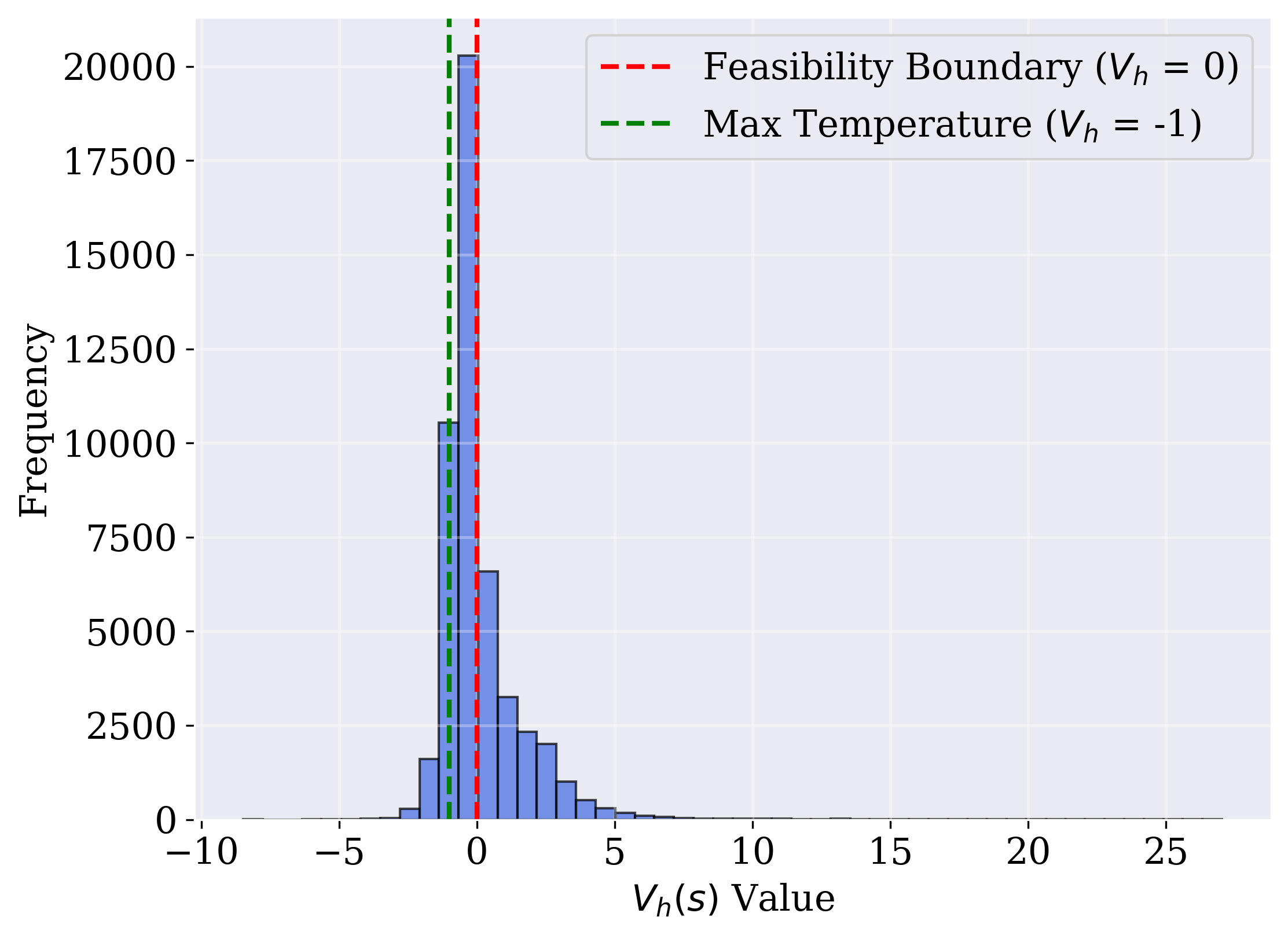}
        \caption{Hopper velocity}
        \label{fig:vh_hopper}
    \end{subfigure}\hfill
    \begin{subfigure}{0.48\textwidth}
        \centering
        \includegraphics[width=\linewidth]{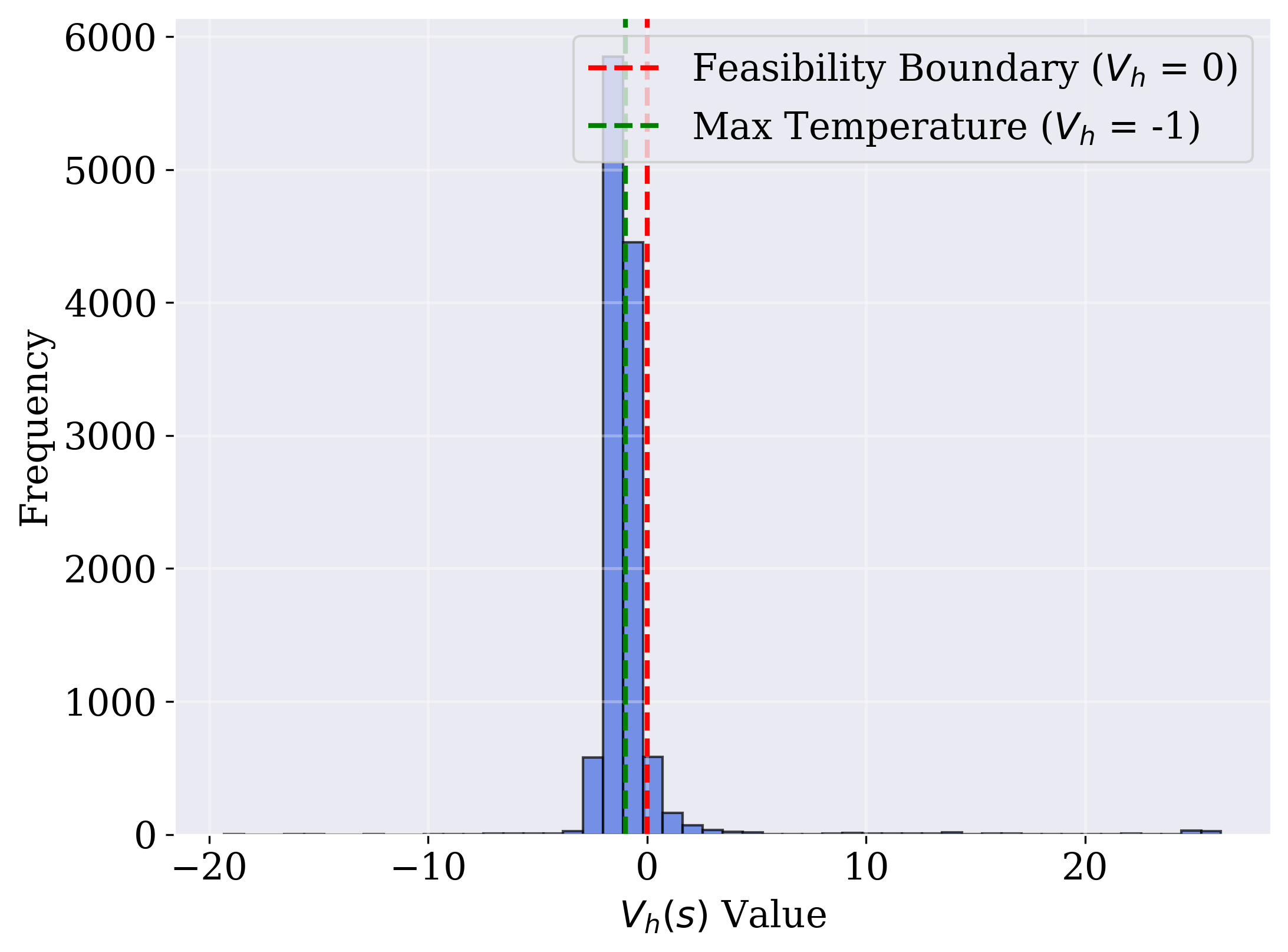}
        \caption{DroneRun}
        \label{fig:vh_drone}
    \end{subfigure}
    \caption{\textbf{Distribution of feasibility values at inference.} Histograms of
    $\hat V_h(s)$ at the states visited during $20$ evaluation episodes. The red dashed line
    denotes the feasibility boundary ($\hat V_h = 0$), and the green dashed line denotes
    $\hat V_h = -1$, below which the latent scale equals its maximum value $\sigma_z(s) = 1$.
    For $\hat V_h(s) \ge -\sigma_{\min} = -0.1$, the latent scale equals $\sigma_{\min}$.}
    \label{fig:vh_distribution}
\end{figure}

Figure~\ref{fig:vh_distribution} shows the resulting distributions. On Hopper, $50.1\%$ of
the visited states satisfy $-1 < \hat V_h(s) < -0.1$ and receive a state-dependent scale,
$10.7\%$ satisfy $\hat V_h(s) \le -1$ and receive the maximum scale $\sigma_z(s) = 1$, and
$39.2\%$ satisfy $\hat V_h(s) \ge -0.1$ and receive the minimum scale $\sigma_{\min}$. On
DroneRun, the corresponding fractions are $39.2\%$, $38.6\%$, and $22.2\%$. The adaptive
scale is therefore state-dependent on a substantial fraction of the visited states, while
$49.9\%$ (Hopper) and $60.8\%$ (DroneRun) of the states receive one of the two clipping
limits. 

\begin{table}[b]
    \centering
    \begin{minipage}[t]{0.56\textwidth}
        \centering
        \caption{\textbf{Effect of the number of candidate actions $N$ at inference.}
        Normalized return and cost of FOCAL for different numbers of decoded candidates.
        Bold: safe configurations ($C \le 1$). Blue: safe configuration with the highest
        return for each task.}
        \label{tab:ablation_sampling}
        \renewcommand{\arraystretch}{1.2}
        \resizebox{\linewidth}{!}{%
        \begin{tabular}{l c c c c c c c c}
            \toprule
            \multirow{2}{*}{\textbf{Task}}
            & \multicolumn{2}{c}{$N = 1$} & \multicolumn{2}{c}{$N = 4$}
            & \multicolumn{2}{c}{$N = 16$} & \multicolumn{2}{c}{$N = 64$} \\
            \cmidrule(lr){2-3} \cmidrule(lr){4-5} \cmidrule(lr){6-7} \cmidrule(lr){8-9}
            & Return $\uparrow$ & Cost $\downarrow$ & Return $\uparrow$ & Cost $\downarrow$
            & Return $\uparrow$ & Cost $\downarrow$ & Return $\uparrow$ & Cost $\downarrow$ \\
            \midrule
            Walker2d & \textbf{\textcolor{blue}{0.75}} & \textbf{\textcolor{blue}{0.50}}
                & \textbf{0.71} & \textbf{0.51} & \textbf{0.72} & \textbf{0.42}
                & \textbf{0.69} & \textbf{0.31} \\
            Hopper & \textbf{0.50} & \textbf{0.40} & \textbf{0.65} & \textbf{0.23}
                & \textbf{\textcolor{blue}{0.83}} & \textbf{\textcolor{blue}{0.00}}
                & \textbf{0.79} & \textbf{0.00} \\
            AntRun & \textbf{0.21} & \textbf{0.00} & \textbf{0.25} & \textbf{0.00}
                & \textbf{\textcolor{blue}{0.34}} & \textbf{\textcolor{blue}{0.00}}
                & \textbf{0.28} & \textbf{0.00} \\
            BallCircle & \textbf{0.52} & \textbf{0.00} & \textbf{0.54} & \textbf{0.00}
                & \textbf{\textcolor{blue}{0.55}} & \textbf{\textcolor{blue}{0.00}}
                & \textbf{0.53} & \textbf{0.00} \\
            BallRun & \textbf{0.24} & \textbf{0.00} & \textbf{0.25} & \textbf{0.00}
                & \textbf{\textcolor{blue}{0.28}} & \textbf{\textcolor{blue}{0.00}}
                & \textbf{0.25} & \textbf{0.00} \\
            \bottomrule
        \end{tabular}%
        }
    \end{minipage}%
    \hfill
    \begin{minipage}[t]{0.41\textwidth}
        \centering
        \caption{\textbf{$\hat Q_h$ classification accuracy.}
FPR: fraction of unsafe steps classified as feasible. FNR: fraction of
safe steps classified as infeasible. Precision: fraction of feasible
classifications not followed by violations. ``--'': undefined.}
        \label{tab:critic_calibration}
        \renewcommand{\arraystretch}{1.2}
        \resizebox{\linewidth}{!}{%
        \begin{tabular}{l c c c c}
            \toprule
            \textbf{Task} & \textbf{Evaluated steps} & \textbf{FPR} $\downarrow$
            & \textbf{FNR} $\downarrow$ & \textbf{Precision} $\uparrow$ \\
            \midrule
            AntVelocity     & 20000 & --    & 0.092 & 1.000 \\
            SwimmerVelocity & 20000 & --    & 0.203 & 1.000 \\
            DroneRun        & 4000  & 0.257 & 0.324 & 0.944 \\
            CarCircle       & 6000  & 0.125 & 0.219 & 0.970 \\
            AntCircle       & 9333  & 0.221 & 0.312 & 0.897 \\
            \bottomrule
        \end{tabular}%
        }
    \end{minipage}
\end{table}

% ---------------------------------------------------------------------------
\subsection{Feasibility Classification Accuracy of the Hazard Critic}
\label{app:critic_calibration}

The constraint satisfaction of FOCAL depends on whether the hazard critic correctly classifies the actions
generated at inference. We evaluate this classification on the actions executed during
evaluation, using the trained policies. At each
step $t$, the predicted label indicates whether the hazard critic classified the executed
action $a_t^{\mathrm{exec}}$ as feasible,
\begin{equation}
\label{eq:pred_label}
\hat y_t =
\begin{cases}
1, & \text{if } \hat Q_h(s_t, a_t^{\mathrm{exec}}) \le 0,\\
0, & \text{otherwise}.
\end{cases}
\end{equation}
The reference label is obtained in hindsight from the maximum cost over the remainder of
the episode, $M_t = \max(c_t, c_{t+1}, \dots, c_T)$:
\begin{equation}
\label{eq:true_label}
y_t =
\begin{cases}
1, & \text{if } M_t \le 0,\\
0, & \text{otherwise}.
\end{cases}
\end{equation}

\begin{table}[t]
    \centering
    \caption{\textbf{Sensitivity to dataset size.} Normalized return and cost (mean $\pm$
    standard deviation over three training runs) of FISOR and FOCAL trained on $25\%$,
    $50\%$, and $100\%$ of the trajectories of the original dataset. Bold: the better of the
    two methods at the same dataset fraction.}
    \label{tab:dataset_size}
    \resizebox{\linewidth}{!}{%
    \begin{tabular}{llcccccc}
        \toprule
        \multirow{2}{*}{\textbf{Task}} & \multirow{2}{*}{\textbf{Metric}}
        & \multicolumn{3}{c}{\textbf{FISOR}} & \multicolumn{3}{c}{\textbf{FOCAL}} \\
        \cmidrule(lr){3-5} \cmidrule(lr){6-8}
        & & \textbf{25\%} & \textbf{50\%} & \textbf{100\%}
          & \textbf{25\%} & \textbf{50\%} & \textbf{100\%} \\
        \midrule
        \multirow{2}{*}{BallCircle}
        & Return $\uparrow$ & $0.29 \pm 0.01$ & $0.34 \pm 0.01$ & $0.34 \pm 0.14$
          & $\mathbf{0.47 \pm 0.00}$ & $\mathbf{0.53 \pm 0.00}$ & $\mathbf{0.55 \pm 0.00}$ \\
        & Cost $\downarrow$ & $0.02 \pm 0.04$ & $\mathbf{0.00 \pm 0.00}$ & $\mathbf{0.00 \pm 0.00}$
          & $\mathbf{0.00 \pm 0.00}$ & $\mathbf{0.00 \pm 0.00}$ & $\mathbf{0.00 \pm 0.00}$ \\
        \midrule
        \multirow{2}{*}{AntRun}
        & Return $\uparrow$ & $\mathbf{0.40 \pm 0.04}$ & $\mathbf{0.44 \pm 0.07}$ & $\mathbf{0.45 \pm 0.03}$
          & $0.26 \pm 0.00$ & $0.34 \pm 0.03$ & $0.34 \pm 0.04$ \\
        & Cost $\downarrow$ & $0.67 \pm 0.19$ & $0.22 \pm 0.03$ & $0.03 \pm 0.00$
          & $\mathbf{0.02 \pm 0.00}$ & $\mathbf{0.18 \pm 0.04}$ & $\mathbf{0.00 \pm 0.00}$ \\
        \midrule
        \multirow{2}{*}{DroneCircle}
        & Return $\uparrow$ & $0.43 \pm 0.01$ & $0.46 \pm 0.01$ & $0.48 \pm 0.12$
          & $\mathbf{0.55 \pm 0.00}$ & $\mathbf{0.51 \pm 0.03}$ & $\mathbf{0.54 \pm 0.02}$ \\
        & Cost $\downarrow$ & $\mathbf{0.07 \pm 0.01}$ & $\mathbf{0.04 \pm 0.02}$ & $\mathbf{0.00 \pm 0.00}$
          & $0.53 \pm 0.17$ & $\mathbf{0.00 \pm 0.00}$ & $\mathbf{0.00 \pm 0.00}$ \\
        \midrule
        \multirow{2}{*}{HalfCheetah}
        & Return $\uparrow$ & $0.83 \pm 0.01$ & $0.89 \pm 0.01$ & $0.89 \pm 0.02$
          & $\mathbf{0.90 \pm 0.04}$ & $\mathbf{0.95 \pm 0.01}$ & $\mathbf{0.95 \pm 0.00}$ \\
        & Cost $\downarrow$ & $\mathbf{0.00 \pm 0.00}$ & $\mathbf{0.00 \pm 0.00}$ & $\mathbf{0.00 \pm 0.00}$
          & $\mathbf{0.00 \pm 0.00}$ & $\mathbf{0.00 \pm 0.00}$ & $\mathbf{0.00 \pm 0.00}$ \\
        \bottomrule
    \end{tabular}%
    }
\end{table}

We call step $t$ safe if $y_t = 1$ and unsafe otherwise, and report three quantities. The
false-positive rate (FPR) is the fraction of unsafe steps at which the executed action was
classified as feasible, this is the error relevant for safety. The false-negative rate
(FNR) is the fraction of safe steps at which the executed action was classified as
infeasible, this error reflects conservatism. The precision is the fraction of steps
classified as feasible that are safe. Since $y_t$ depends on the actions taken by FOCAL
after step $t$, these errors reflect both the hazard critic and the subsequent actions of
the policy.

Table~\ref{tab:critic_calibration} reports the results. On DroneRun, CarCircle, and
AntCircle, the FPR ranges from $0.125$ to $0.257$, that is, between $12.5\%$ and $25.7\%$ of
the steps followed by a violation were executed with an action classified as feasible. On
AntVelocity and SwimmerVelocity, no violations occurred during evaluation, so the FPR is
undefined. The FNR ranges from $0.092$ to $0.324$ across tasks.
These results show that the hazard critic misclassifies a non-negligible fraction of
unsafe actions as safe.

\begin{table}[b]
    \centering
    \caption{\textbf{Training with limited coverage of the constraint boundary.} For each
    task, FOCAL is trained on the $20\%$ of trajectories with the lowest cumulative cost.
    Cum.\ cost: average undiscounted episodic cost of the retained trajectories. Return and
    cost: normalized, mean $\pm$ standard deviation over three training runs.}
    \label{tab:stress_test}
    \begin{tabular}{l c c c c}
        \toprule
        \textbf{Task} & \textbf{Retained trajectories} & \textbf{Cum.\ cost}
        & \textbf{Return} $\uparrow$ & \textbf{Cost} $\downarrow$ \\
        \midrule
        AntVelocity    & 449 (20.0\%) & 10.46 & $0.46 \pm 0.15$ & $0.00 \pm 0.00$ \\
        Hopper         & 448 (20.0\%) & 18.87 & $0.27 \pm 0.10$ & $0.00 \pm 0.00$ \\
        HalfCheetah    & 499 (20.0\%) & 10.79 & $0.94 \pm 0.01$ & $0.00 \pm 0.00$ \\
        \bottomrule
    \end{tabular}
\end{table}
% ---------------------------------------------------------------------------
\subsection{Sensitivity to Dataset Size}
\label{app:dataset_size}

To evaluate the sensitivity of FOCAL to the amount of offline data, we subsample the
trajectories of each dataset to $25\%$, $50\%$, and $100\%$ of the original DSRL dataset and
train FOCAL and FISOR on each subset. Table~\ref{tab:dataset_size} reports the mean and
standard deviation over three independent training runs.

Both methods satisfy the cost threshold at all dataset fractions on the evaluated tasks.
Reducing the dataset size generally decreases the normalized return of both methods, while
its effect on the normalized cost depends on the task. FOCAL obtains a higher return than
FISOR on most of the evaluated tasks across dataset fractions. These results indicate that
FOCAL remains within the cost threshold when the amount of offline data is reduced, at the
cost of a lower return.

% ---------------------------------------------------------------------------
\subsection{Training with Limited Coverage of the Constraint Boundary}
\label{app:stress_test}

FOCAL learns the feasibility critics from the transitions in the dataset, including
transitions that violate the constraint. To evaluate its behavior when the dataset contains
few such transitions, we construct datasets with reduced coverage of the constraint
boundary. Since trajectories with zero cumulative cost constitute less than $3\%$ of the
respective DSRL datasets, training only on such trajectories is not feasible. We instead
retain, for each task, the $20\%$ of trajectories with the lowest cumulative cost, retrain
FOCAL on the resulting dataset, and evaluate it with the protocol of
Section~\ref{sec:setup}.

Table~\ref{tab:stress_test} reports the retained datasets and the results. FOCAL remains
within the cost threshold on all evaluated tasks, while its normalized return is generally
lower than when trained on the full dataset. The retained trajectories still contain some
constraint violations, so this experiment evaluates reduced, rather than absent, coverage of
the constraint boundary.

\begin{algorithm}[t]
\caption{Feasibility-Optimized Conditional Actor Learning (FOCAL)}
\label{alg:focal}
\begin{algorithmic}[1]
\Require Offline dataset $\mathcal{D}$, number of gradient steps $K$, warm-up steps
$K_{\mathrm{warm}}$, number of candidates $N$, minimum latent scale $\sigma_{\min}$
\State Initialize $\hat V_r$, $\hat V_h$, twin Q-networks $\{\hat Q_{r,i}\}_{i=1,2}$ and
$\{\hat Q_{h,i}\}_{i=1,2}$, their target copies, encoder $q_\phi$, and decoder $D_\theta$
\Statex \textbf{\textit{Training}}
\For{$k = 1, \dots, K$}
    \State Sample a mini-batch $\mathcal{B} \sim \mathcal{D}$
    \State Update $\hat V_r$ and $\hat Q_{r,i}$ with Eqs.~(\ref{eq:vr_loss})--(\ref{eq:qr_loss})
    \State Update $\hat V_h$ and $\hat Q_{h,i}$ with Eqs.~(\ref{eq:vh_loss})--(\ref{eq:qh_loss})
    \If{$k > K_{\mathrm{warm}}$}
        \State Compute the masks (Eq.~(\ref{eq:masks})) and step sizes
        (Eq.~(\ref{eq:step_sizes})) on $\mathcal{B}$
        \State Update $\phi$ and $\theta$ by minimizing $\mathcal{L}_{\mathrm{CVAE}}$
        (Eq.~(\ref{eq:cvae_objective})) with the critics held fixed
    \EndIf
    \State Update the target Q-networks by Polyak averaging
\EndFor
\Statex \textbf{\textit{Inference at state $s$}}
\State Compute $\sigma_z(s) = \mathrm{clip}(-\hat V_h(s), \sigma_{\min}, 1)$
(Eq.~(\ref{eq:adaptive_variance}))
\State Sample $z_i \sim \mathcal{N}(0, \sigma_z(s)^2 I)$ and decode
$\hat a_i = D_\theta(s, z_i)$ for $i = 1, \dots, N$
\State $\mathcal{C}(s) \gets \{\, i : \hat Q_h(s, \hat a_i) \le 0 \,\}$
\If{$\mathcal{C}(s) \neq \emptyset$}
    \State Execute $\hat a_{i^\star}$ with $i^\star = \arg\max_{i \in \mathcal{C}(s)} \hat Q_r(s, \hat a_i)$
\Else
    \State Execute $\hat a_{i^\star}$ with $i^\star = \arg\min_{i \in \{1,\dots,N\}} \hat Q_h(s, \hat a_i)$
\EndIf
\end{algorithmic}
\end{algorithm}

% Requires: algorithm, algpseudocode (algorithmicx), graphicx, subcaption, booktabs, xcolor.
\providecommand{\todo}[1]{\textcolor{red}{[TODO: #1]}}

\section{Experimental Details}
\label{app:experimental_details}

This section describes the training procedure and hyperparameters of FOCAL, the baseline
implementations, and the benchmark tasks.

% ---------------------------------------------------------------------------
\subsection{Training Procedure and Implementation}
\label{app:implementation_details}

Our implementation builds on the official implementation of LSPC~\cite{koirala2024latent}.
Training consists of $10^6$ gradient steps. During the first $K_{\mathrm{warm}} = 10^5$
steps, only the reward and feasibility critics are trained. During the remaining
$9 \times 10^5$ steps, the critics and the CVAE actor are updated at every step. In each
actor update, the critics are treated as constants, so that the actor gradient does not
affect critic training. The warm-up phase ensures that the feasibility masks and the
first-order terms in Eq.~(\ref{eq:cvae_objective}) are computed from critics that have been
trained before the actor is updated. All expectations are approximated with mini-batches
sampled uniformly from the offline dataset. Following the implementation of LSPC~\cite{koirala2024latent}, the scaled latent samples are additionally clipped elementwise to $[-0.5, 0.5]$ before decoding. Target copies of the twin Q-networks are
maintained and updated by Polyak averaging with rate $\rho$. The normalized step sizes
$\lambda_r$ and $\lambda_h$ (Eq.~(\ref{eq:step_sizes})) are computed on each mini-batch.
Algorithm~\ref{alg:focal} summarizes training and inference, and
Table~\ref{tab:hyperparameters} lists the hyperparameters, which are shared across all
tasks. 

\begin{table}[t]
    \centering
    \caption{\textbf{Hyperparameters of FOCAL.} The same values are used for all tasks.}
    \label{tab:hyperparameters}
    \renewcommand{\arraystretch}{1.15}
    \begin{tabular}{l c}
        \toprule
        \textbf{Hyperparameter} & \textbf{Value} \\
        \midrule
        \multicolumn{2}{c}{\textit{Networks and optimization}} \\
        \midrule
        Hidden layers & 2 \\
        Hidden dimension & 256 \\
        Activation function & ReLU \\
        Latent dimension of $z$ & 32 \\
        Optimizer & Adam \\
        Learning rates (critics / actor) & 3e-4 \\
        Mini-batch size & 1024 \\
        Total gradient steps $K$ & $10^6$ \\
        Discount factor $\gamma$ & 0.99 \\
        Target network update rate $\rho$ & 0.005 \\
        \midrule
        \multicolumn{2}{c}{\textit{Critic learning}} \\
        \midrule
        Reward expectile $\omega_r$ & 0.7 \\
        Hazard expectile $\omega_h$ & 0.9 \\
        Constraint violation values $h(s)$ (safe / unsafe) & $-1$ / $M = 25$ \\
        Critic warm-up steps $K_{\mathrm{warm}}$ & $10^5$ \\
        \midrule
        \multicolumn{2}{c}{\textit{Actor learning}} \\
        \midrule
        KL weight $\beta$ & 0.5 \\
        Step-size constants $\alpha_r, \alpha_h$ & 2.5 \\
        \midrule
        \multicolumn{2}{c}{\textit{Inference}} \\
        \midrule
        Number of candidates $N$ & 16 \\
        Minimum latent scale $\sigma_{\min}$ & 0.1 \\
        \bottomrule
    \end{tabular}
\end{table}

\paragraph{Baselines.}
% TODO: confirm which baseline results are taken from Zheng et al. (2024) and which were rerun.
For BC, CDT, BCQ-Lag, CPQ, and COptiDICE, we use the implementations and configurations of
the OSRL library~\cite{liu2023datasets}. For FISOR~\cite{zheng2024safe} and
LSPC~\cite{koirala2024latent}, we use the official implementations with the
hyperparameters reported for the DSRL benchmark.

\paragraph{Compute resources.}
All experiments were conducted on a single NVIDIA A100 GPU. Training FOCAL for one seed on
one task, including the critic warm-up phase, takes approximately 3 hours.
% ---------------------------------------------------------------------------
\subsection{Benchmark Tasks}
\label{app:benchmark_details}

\begin{figure}[t]
    \centering
    \begin{subfigure}[b]{0.32\textwidth}
        \centering
        \includegraphics[width=\textwidth, height=3.0cm]{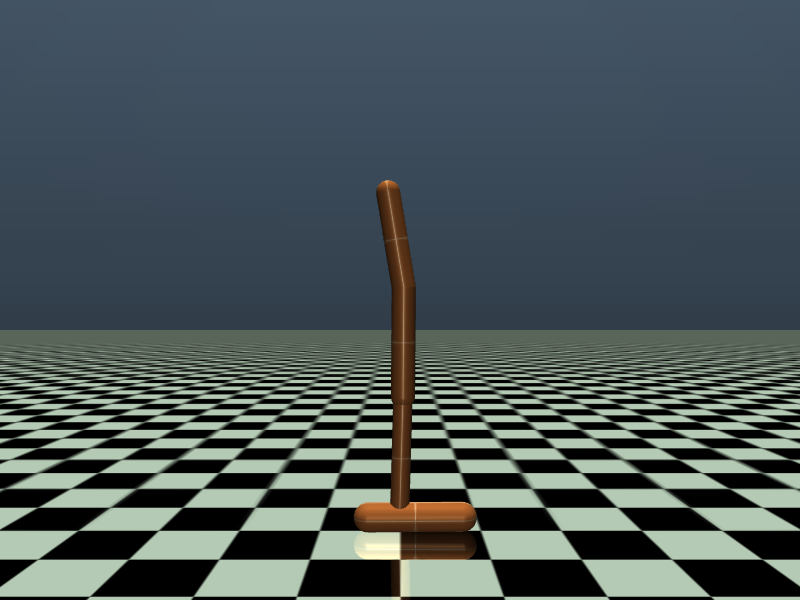}
        \caption{HopperVelocity}
        \label{fig:safety_gym1}
    \end{subfigure}
    \hfill
    \begin{subfigure}[b]{0.32\textwidth}
        \centering
        \includegraphics[width=\textwidth, height=3.0cm]{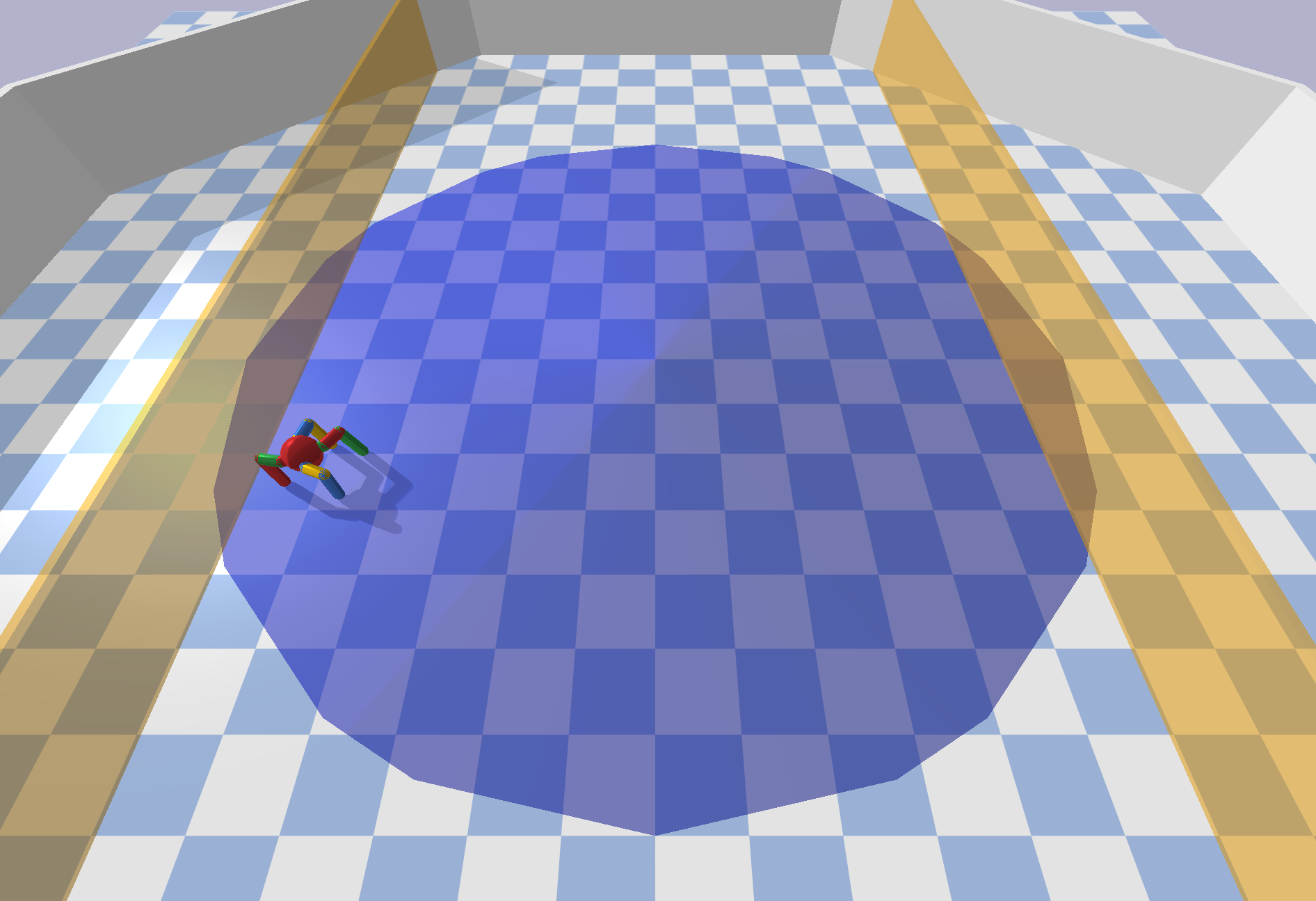}
        \caption{AntCircle}
        \label{fig:bullet_gym1}
    \end{subfigure}
    \hfill
    \begin{subfigure}[b]{0.32\textwidth}
        \centering
        \includegraphics[width=\textwidth, height=3.0cm]{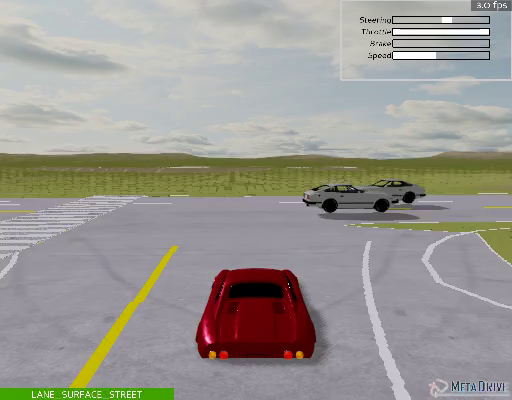}
        \caption{MetaDrive}
        \label{fig:env_md1}
    \end{subfigure}

    \vspace{4mm}

    \begin{subfigure}[b]{0.32\textwidth}
        \centering
        \includegraphics[width=\textwidth, height=3.0cm]{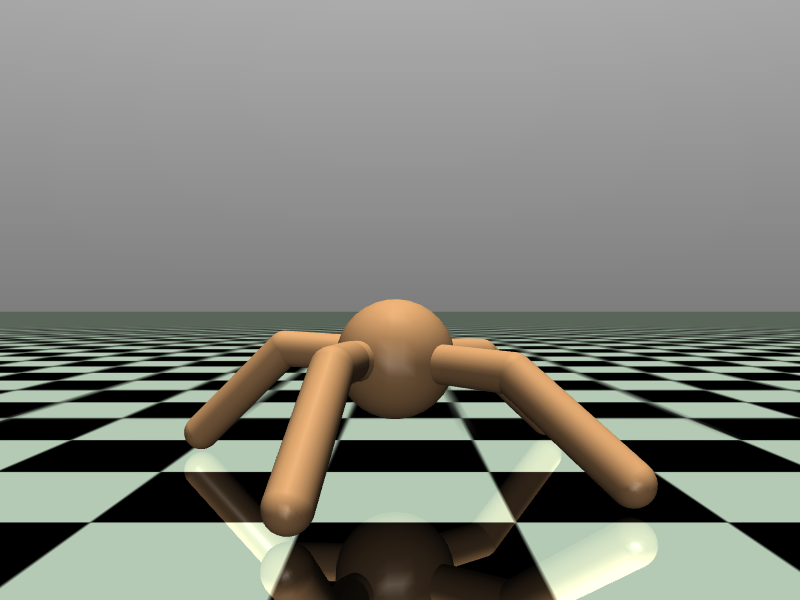}
        \caption{AntVelocity}
        \label{fig:safety_gym2}
    \end{subfigure}
    \hfill
    \begin{subfigure}[b]{0.32\textwidth}
        \centering
        \includegraphics[width=\textwidth, height=3.0cm]{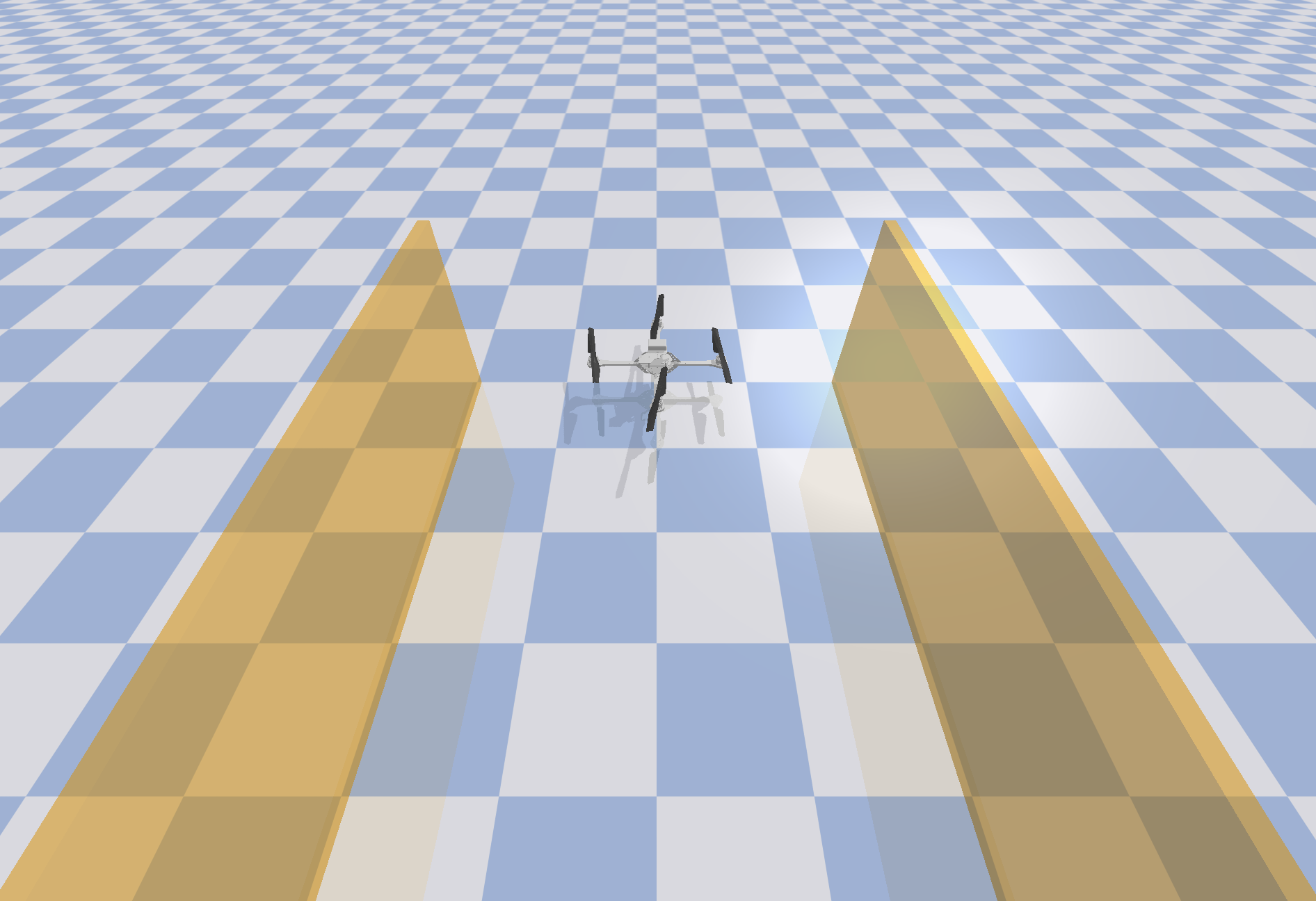}
        \caption{DroneRun}
        \label{fig:bullet_gym2}
    \end{subfigure}
    \hfill
    \begin{subfigure}[b]{0.32\textwidth}
        \centering
        \includegraphics[width=\textwidth, height=3.0cm]{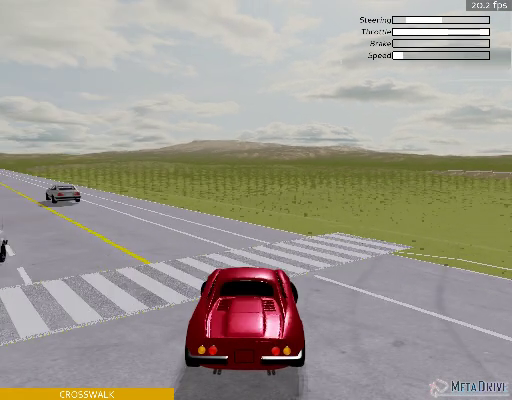}
        \caption{MetaDrive}
        \label{fig:env_md2}
    \end{subfigure}
    \caption{\textbf{Representative DSRL tasks.} Left column: Safety-Gymnasium. Middle
    column: Bullet-Safety-Gym. Right column: MetaDrive.}
    \label{fig:all_envs}
\end{figure}

All experiments use the Datasets for Safe Reinforcement Learning (DSRL)
benchmark~\cite{liu2023datasets}, which provides offline datasets for safe RL that contain
both safe and unsafe trajectories with varying returns and costs.
Figure~\ref{fig:all_envs} shows representative tasks.

\paragraph{Safety-Gymnasium.}
Safety-Gymnasium~\cite{ji2023safety} is a MuJoCo-based benchmark. We use two types of
tasks. In the navigation tasks, a car-like agent must
complete an objective while
avoiding hazard regions. In the velocity tasks, the agent is rewarded for moving
forward and incurs a cost when its velocity exceeds a threshold.

\paragraph{Bullet-Safety-Gym.}
Bullet-Safety-Gym~\cite{gronauer2022bullet} is built on the PyBullet engine and uses four
agents (Ball, Car, Drone, and Ant) with different dynamics. In the Run tasks, the agent is
rewarded for moving forward and incurs a cost when it leaves a safe region or exceeds a
velocity threshold. In the Circle tasks, the agent is rewarded for moving along a circle
and incurs a cost when it leaves a safe region. The episodes are shorter than in
Safety-Gymnasium.

\paragraph{MetaDrive.}
MetaDrive~\cite{li2022metadrive} is an autonomous driving simulator. The agent must reach a
destination while avoiding collisions with other vehicles and leaving the road. The nine
tasks combine three levels of road difficulty (Easy, Medium, Hard) with three levels of
traffic density (Sparse, Mean, Dense).

\newpage
\section*{NeurIPS Paper Checklist}

\begin{enumerate}

\item {\bf Claims}
    \item[] Question: Do the main claims made in the abstract and introduction accurately reflect the paper's contributions and scope?
    \item[] Answer: \answerYes{} % Replace by \answerYes{}, \answerNo{}, or \answerNA{}.
    \item[] Justification: Please refer to Section \ref{sec:method} and \ref{sec:results}.

\item {\bf Limitations}
    \item[] Question: Does the paper discuss the limitations of the work performed by the authors?
    \item[] Answer: \answerYes{} % Replace by \answerYes{}, \answerNo{}, or \answerNA{}.
    \item[] Justification: Please refer to Section \ref{sec:conclusion}.

\item {\bf Theory assumptions and proofs}
    \item[] Question: For each theoretical result, does the paper provide the full set of assumptions and a complete (and correct) proof?
    \item[] Answer: \answerYes{} % Replace by \answerYes{}, \answerNo{}, or \answerNA{}.
    \item[] Justification: Please refer to Section \ref{sec:method} and \ref{app:theoretical_proofs}.

    \item {\bf Experimental result reproducibility}
    \item[] Question: Does the paper fully disclose all the information needed to reproduce the main experimental results of the paper to the extent that it affects the main claims and/or conclusions of the paper (regardless of whether the code and data are provided or not)?
    \item[] Answer: \answerYes{} % Replace by \answerYes{}, \answerNo{}, or \answerNA{}.
    \item[] Justification: Please refer Section \ref{sec:method} and \ref{app:experimental_details}.

\item {\bf Open access to data and code}
    \item[] Question: Does the paper provide open access to the data and code, with sufficient instructions to faithfully reproduce the main experimental results, as described in supplemental material?
    \item[] Answer: \answerYes{} % Replace by \answerYes{}, \answerNo{}, or \answerNA{}.
    \item[] Justification: The complete code for the paper can be found at this \href{https://github.com/mahesh-keswani/FOCAL-Offline-SafeRL}{link}.

\item {\bf Experimental setting/details}
    \item[] Question: Does the paper specify all the training and test details (e.g., data splits, hyperparameters, how they were chosen, type of optimizer) necessary to understand the results?
    \item[] Answer: \answerYes{} % Replace by \answerYes{}, \answerNo{}, or \answerNA{}.
    \item[] Justification: Please refer to Section \ref{app:experimental_details}.

\item {\bf Experiment statistical significance}
    \item[] Question: Does the paper report error bars suitably and correctly defined or other appropriate information about the statistical significance of the experiments?
    \item[] Answer: \answerYes{} % Replace by \answerYes{}, \answerNo{}, or \answerNA{}.
    \item[] Justification: Please refer to Section \ref{sec:results}.

\item {\bf Experiments compute resources}
    \item[] Question: For each experiment, does the paper provide sufficient information on the computer resources (type of compute workers, memory, time of execution) needed to reproduce the experiments?
    \item[] Answer: \answerYes{} % Replace by \answerYes{}, \answerNo{}, or \answerNA{}.
    \item[] Justification: Please refer to Section \ref{app:experimental_details}.
    
\item {\bf Code of ethics}
    \item[] Question: Does the research conducted in the paper conform, in every respect, with the NeurIPS Code of Ethics \url{https://neurips.cc/public/EthicsGuidelines}?
    \item[] Answer: \answerYes{} % Replace by \answerYes{}, \answerNo{}, or \answerNA{}.
    \item[] Justification: The paper conforms to the NeurIPS Code of Ethics in all respects.

\item {\bf Broader impacts}
    \item[] Question: Does the paper discuss both potential positive societal impacts and negative societal impacts of the work performed?
    \item[] Answer: \answerNA{} % Replace by \answerYes{}, \answerNo{}, or \answerNA{}.
    \item[] Justification: This work focuses on foundational algorithmic research in safe offline RL evaluated solely in simulated environments. It does not involve real-world deployment or direct applications that present immediate negative societal impacts. Therefore, a dedicated discussion on broader societal impacts was not included.
    
\item {\bf Safeguards}
    \item[] Question: Does the paper describe safeguards that have been put in place for responsible release of data or models that have a high risk for misuse (e.g., pre-trained language models, image generators, or scraped datasets)?
    \item[] Answer: \answerNA{} % Replace by \answerYes{}, \answerNo{}, or \answerNA{}.
    \item[] Justification: The paper poses no such risks.

\item {\bf Licenses for existing assets}
    \item[] Question: Are the creators or original owners of assets (e.g., code, data, models), used in the paper, properly credited and are the license and terms of use explicitly mentioned and properly respected?
    \item[] Answer: \answerYes{} % Replace by \answerYes{}, \answerNo{}, or \answerNA{}.
    \item[] Justification: The paper properly cite all external code used in our research and adhere to their license requirements.

\item {\bf New assets}
    \item[] Question: Are new assets introduced in the paper well documented and is the documentation provided alongside the assets?
    \item[] Answer: \answerYes{} % Replace by \answerYes{}, \answerNo{}, or \answerNA{}.
    \item[] Justification: The code for FOCAL is documented, with setup, training instructions provided in the README of the public \href{https://github.com/mahesh-keswani/FOCAL-Offline-SafeRL}{repository}.

\item {\bf Crowdsourcing and research with human subjects}
    \item[] Question: For crowdsourcing experiments and research with human subjects, does the paper include the full text of instructions given to participants and screenshots, if applicable, as well as details about compensation (if any)? 
    \item[] Answer: \answerNA{} % Replace by \answerYes{}, \answerNo{}, or \answerNA{}.
    \item[] Justification: The paper does not involve crowdsourcing nor research with human subjects.

\item {\bf Institutional review board (IRB) approvals or equivalent for research with human subjects}
    \item[] Question: Does the paper describe potential risks incurred by study participants, whether such risks were disclosed to the subjects, and whether Institutional Review Board (IRB) approvals (or an equivalent approval/review based on the requirements of your country or institution) were obtained?
    \item[] Answer: \answerNA{} % Replace by \answerYes{}, \answerNo{}, or \answerNA{}.
    \item[] Justification: The paper does not involve crowdsourcing nor research with human subjects.

\item {\bf Declaration of LLM usage}
    \item[] Question: Does the paper describe the usage of LLMs if it is an important, original, or non-standard component of the core methods in this research? Note that if the LLM is used only for writing, editing, or formatting purposes and does \emph{not} impact the core methodology, scientific rigor, or originality of the research, declaration is not required.
    %this research? 
    \item[] Answer: \answerNA{} % Replace by \answerYes{}, \answerNo{}, or \answerNA{}.
    \item[] Justification: Core method development in this research does not involve LLMs as any important, original, or non-standard components.

\end{enumerate}

\end{document}